\documentclass[letterpaper]{article} 

\newif\ifdraftmode
\draftmodetrue 

\usepackage[dvipsnames]{xcolor} 
\definecolor{BrickRed}{rgb}{0.71, 0.09, 0.25}

\usepackage[size=scriptsize]{todonotes}

\usepackage[preprint]{aaai2027}
\usepackage[hyphens]{url}  
\usepackage{graphicx} 
\usepackage{natbib}  
\usepackage{caption} 
\usepackage{booktabs}
\usepackage{multirow}
\usepackage{rotating}        
\usepackage{threeparttable}  

\usepackage{algorithm}
\usepackage{algpseudocode}

\usepackage{amssymb}     
\usepackage{amsmath}
\usepackage{mathtools}
\usepackage{physics}
\usepackage{comment}
\usepackage{soul}             
\newtheorem{theorem}{Theorem}[section]

\newtheorem{proposition}{Proposition}[section]

\newtheorem{remark}{Remark}[section]

\usepackage[acronym]{glossaries}

\newacronym{acr:bai}{BAI}{Best Arm Identification}
\newacronym{acr:shrr}{SH-RR}{Sequential Halving with Resource Rationing}
\newacronym{acr:sh}{SH}{Sequential Halving}
\newacronym{acr:shprr}{\mbox{SHPRR}}{Sequential Halving with Proportional Regret Rationing}
\newacronym{acr:tt}{TTSPRT}{Top-Two SPRT}
\newacronym{acr:sp}{SP}{Success Probability}
\newacronym{acr:nsr}{NSR}{Normalized Simple Regret}

\usepackage{adjustbox}   
\usepackage{import}
\usepackage{subcaption}  
\usepackage{relsize}
\usepackage{pgfplots}
\pgfplotsset{compat=1.18} 
\usepgfplotslibrary{groupplots}

\usepackage{tikz}
\usetikzlibrary{calc,fit,positioning,shapes.geometric,arrows.meta}

\title{SHPRR: An Algorithm for Pure Exploration under Constrained Regret}

\author{
    Georgina Nouli\textsuperscript{\rm 1},
    Flore Sentenac\textsuperscript{\rm 2},
    Maximilian Schiffer\textsuperscript{\rm 2}
}

\affiliations{
    \textsuperscript{\rm 1}School of Management, Technical University of Munich, Munich, Germany\\
    \textsuperscript{\rm 2} HEC Paris, Jouy-en-Josas, France
}

\begin{document}

\maketitle

\begin{abstract}
Best Arm Identification (BAI) conventionally measures the cost of experimentation by the number of samples collected. In many operational settings, however, the cost of exploration relative to a trusted baseline already in place is equally important. We study BAI under both a fixed sampling horizon and a regret budget. Unlike generic resources considered in constrained pure exploration, regret has a distinctive structure: sampling better arms is less costly relative to the baseline. We theoretically show that the regret cost of exploration decreases across the phases of Sequential Halving, a widely used pure-exploration algorithm. This observation yields improved guarantees on the probability of identifying the best arm and motivates a more effective allocation of the regret budget. We propose Sequential Halving with Proportional Regret Rationing (SHPRR), which front-loads the regret budget to allow more exploration when many arms remain. Experiments on synthetic and real-world instances show that SHPRR improves best-arm identification performance over uniform regret allocation and existing baselines.
\end{abstract}

\section{Introduction}
Pure exploration in multi-armed bandits, formalized as \gls{acr:bai}, has established foundational paradigms for sample-efficient decision-making under uncertainty~\citep{kalyanakrishnan2012pac,garivier2016optimal,audibert2010best,karnin2013almost}. In classical settings—such as offline algorithm selection or clinical pre-screening—the cost of learning is measured naturally in \emph{sample complexity}: every pull incurs an identical, uniform cost, and the goal is to identify the optimal arm using as few observations as possible. 

However, in many live deployment settings, experimentation does not occur in a vacuum. Instead, it runs directly against a trusted, operational baseline, e.g., a production recommender or a standard-of-care treatment. Here, sampling an inferior candidate inflicts an immediate, tangible penalty on the live system. While sample-based bounds remain invaluable for offline evaluation, live deployments demand an explicit tolerance for cumulative loss. An online A/B test, for instance, cannot simply run for a fixed sample quota if cumulative degradation against the production system exceeds a strict threshold. Accordingly, regret evolves from a post-hoc evaluation metric into a scarce, active budget constraint that directly governs the exploration process in such settings.

While \gls{acr:bai} with resource constraints has been studied \citep{li2024best}, budgeting realized regret introduces two structural challenges that set it apart from standard exogenous cost models (e.g., financial or computational limits): i) \emph{Signed consumption:} outperforming the baseline yields a negative cost, effectively replenishing the remaining budget rather than draining it. ii) \emph{Consumption--reward coupling:} consumption is not an externally sampled
attribute but the recentered reward process itself. Hence, better options are less costly, a phenomenon that can be leveraged in the exploration.

\gls{acr:sh} \citep{karnin2013almost} provides a natural foundation for phase-based exploration under resource constraints. However, uniform regret allocations across phases fail to exploit a critical structural dynamic of elimination-based algorithms: as suboptimal arms are pruned, the average suboptimality gap of the surviving set contracts. Because early phases must sample from a broader, unrefined pool of arms, the expected regret cost per pull is highest at the start of the algorithm and steadily decreases across phases. 

This observation directly motivates our proposed algorithm, \gls{acr:shprr}. Rather than distributing the regret budget uniformly, \gls{acr:shprr} strategically front-loads the budget toward earlier phases, where information acquisition incurs the highest regret in proportion to the number of surviving arms. Controlled by a single tuning parameter $\alpha \in [0,1]$, \gls{acr:shprr} smoothly interpolates between classical uniform allocation ($\alpha = 0$) and aggressive early-phase budget prioritization ($\alpha = 1$), while guaranteeing budget enforcement. Across extensive synthetic and real-world benchmarks, \gls{acr:shprr} consistently matches or outperforms \gls{acr:shrr}, AT-LUCB, and \gls{acr:tt}. Crucially, while competing baselines demonstrate brittle performance, \gls{acr:shprr} remains robust across all settings.

Our contributions are summarized as follows: we formalize \gls{acr:bai} under a realized regret budget, capturing real-world deployment settings where cumulative underperformance against a baseline—rather than sample count—is the critical resource. We characterize two fundamental structural properties that distinguish regret budgeting from standard resource constraints: \emph{signed consumption} and \emph{consumption--reward coupling}. Furthermore, we reveal that elimination-based algorithms exhibit a contracting average suboptimality gap across phases, establishing that early-phase exploration is inherently the most expensive in terms of regret.

Guided by this phase-contraction insight, we propose \gls{acr:shprr}, which strategically front-loads the regret budget toward early phases in proportion to the active arm count, controlled by a single parameter $\alpha \in [0,1]$.

We provide theoretical guarantees showing that \gls{acr:shprr} controls failure probability under bounded regret. Empirically, \gls{acr:shprr} consistently matches or outperforms \gls{acr:shrr}, AT-LUCB, and \gls{acr:tt} across synthetic and real-world environments. Crucially, while existing baselines display brittle behavior, performing well in specific settings but collapsing in others, \gls{acr:shprr} maintains strong, robust performance across all domains.

\section{Related Work}

Literature on \gls{acr:bai} follows one of two formulations: fixed-confidence (PAC) \gls{acr:bai}, where the algorithm must identify the best arm with probability at least $1-\delta$ while minimizing pulls \cite{kalyanakrishnan2012pac, garivier2016optimal, kaufmann2016complexity}, and fixed-budget \gls{acr:bai}, where a limited budget $T$ is used to maximize the probability of correct identification \cite{audibert2010best, karnin2013almost, gabillon2012best, carpentier2016tight}. Identification quality is  measured via \gls{acr:sp}, or via simple regret -- the expected gap between the best arm's mean and that of the recommended arm \cite{bubeck2009pure, bubeck2011pure, zhao2023revisiting}.

\textit{Elimination and Confidence-Bound Methods:} \gls{acr:sh} \cite{karnin2013almost} and Successive Rejects \cite{audibert2010best} allocate an increasing number of pulls per arm across rounds while discarding the weakest arms. \citet{carpentier2016tight} established a matching lower bound for the latter, closing the gap with existing upper bounds, while \citet{barrier2023best} study fixed-budget \gls{acr:bai} in non-parametric models, sharpening these bounds via KL-based large deviation arguments. Confidence-bound methods instead use interval estimates to guide sampling and stopping: LUCB \cite{kalyanakrishnan2012pac} pulls the empirical best arm together with its most optimistic competitor, KL-LUCB \cite{kaufmann2013information} and lil'UCB \cite{jamieson2014lil} sharpen the underlying confidence intervals, and Track-and-Stop \cite{garivier2016optimal} instead tracks the sampling proportions prescribed by the instance-dependent lower bound, attaining asymptotically optimal sample complexity. \citet{jun2016anytime} extended LUCB to the anytime setting (AT-LUCB), while UGapE \cite{gabillon2012best} unifies the confidence-bound principle across the fixed-budget and fixed-confidence settings.

\textit{Top-Two Methods:} \citet{russo2016simple} introduced Top-Two Thompson Sampling (TTTS), which maintains a posterior over each arm's mean and performs randomized sampling 
between the current best arm and a challenger; \citet{shang2020fixed} and \citet{jourdan2022toptwo} later established $\delta$-PAC guarantees and sample complexity bounds for TTTS variants. \citet{mukherjee2023sprt} adopt a related top-two structure but replace posterior sampling with Sequential Probability Ratio Test (SPRT) likelihood-ratio statistics to select arms and decide when to stop, yielding a $\delta$-PAC algorithm that is asymptotically optimal for exponential-family rewards without the computational cost of posterior sampling. None of these classical formulations account for the cost of arm pulls, feasibility constraints, or the regret incurred while exploring, motivating the lines of work surveyed next.

\textit{\gls{acr:bai} with Costs, Constraints and Cumulative Regret:} \gls{acr:bai} with safety constraints imposes per-pull feasibility requirements \cite{wang2022best, sui2015safe}, while variance-constrained \gls{acr:bai} \cite{hou2022almost} instead constrains the variance of the identified arm. \citet{poiani2022multifidelity} and \citet{wang2023multi} study multi-fidelity \gls{acr:bai}, where biased approximations of each arm's reward are available at varying known costs, and the goal is to minimize total identification cost while guaranteeing correct identification with high probability. \citet{kanarios2024cost} study Cost-Aware \gls{acr:bai} (CABAI), identifying the highest-reward arm at minimum expected cost, whereas \citet{qin2025cost} target the highest reward-to-cost ratio using the fewest samples. \citet{lardy2026constrained} study BAI constrained to the subset of arms with costs under a specific threshold when reward and cost are dependent. \citet{harding2026balancing} minimize a risk functional jointly penalizing a performance criterion and cumulative sampling cost, avoiding the need to pre-specify a budget or confidence level and \citet{yang2024best} couple identification quality with the cumulative regret incurred during exploration.
The closest setting to ours is \citet{li2024best}, who impose cumulative resource-consumption constraints across arm pulls, and develop \gls{acr:shrr}, an extension of SH that eliminates arms in phases while uniformly rationing each phase's budget given arm costs. Our setting differs from \citet{li2024best} in two respects. First, \emph{consumption is signed}: because it is an arm's regret relative to a baseline, outperforming the baseline yields a negative cost that replenishes the budget rather than draining it, unlike the strictly non-negative costs assumed there. Second, \emph{consumption is coupled to the reward process}: it is not an externally sampled attribute but the recentered reward itself, rendering better arms cheaper to explore.

\section{Problem Setting}

We consider a stochastic multi-armed bandit problem augmented with an \emph{exogenous baseline} and a \emph{realized regret budget}, in addition to the usual sampling horizon.

\subsection{Bandit Model}

Let $ S=\{1,\dots,K\}$ be a finite set of $K$ arms available to the agent. Pulling arm $i\in S$ at round $t$ yields a stochastic reward $r_t\sim\nu_i$, bounded almost surely in $[0,1]$, with mean $\mu_i=\mathbb{E}[r_t]$. We assume a unique arm with highest mean reward, and wlog that $\mu_1>\mu_2\geq \ldots \geq \mu_K$. We denote by $\mu^\star=\mu_{1}$ the mean of the optimal arm, and write $\Delta_i=\mu^\star-\mu_i\ge 0$ for the suboptimality gap of arm $i$. The distributions $\nu_i$, and in particular the means $\mu_i$, are unknown to the agent for every $i\in S$ at the beginning of the run.

In addition to $S$, there is a \emph{baseline arm} $k_0\notin S$, representing an incumbent policy or status-quo action against which arms in $S$ are evaluated. Unlike the arms in $S$, the baseline arm is not available to the agent: it is never pulled, and its reward distribution $\nu_0$ is never observed. Only its mean $\mu_0=\mathbb{E}_{\nu_0}[r]$ is known, provided as a fixed reference point. We assume $\mu_0\le\mu^\star$, i.e., the baseline is weakly dominated by the best available arm, reflecting its role as an incumbent reference rather than an unattainable ideal.

A policy $\pi$ sequentially selects arms, with $k_t$ being the arm selected at time $t$, and, at stopping time $\tau$, outputs a recommendation $\phi\in S$. Formally, let $\mathcal{F}_t=\sigma(k_1,r_1,\dots,k_t,r_t)$ be the filtration generated by the history of pulls and observed rewards, and let $\tau$ be a stopping time with respect to $(\mathcal{F}_t)_{t\ge 0}$. At each round $t\le\tau$, $\pi$ selects~$k_t$ as a function of $\mathcal{F}_{t-1}$; at $\tau$, $\pi$ outputs $\phi$ as a function of~$\mathcal{F}_\tau$.

\paragraph{Realized baseline regret.}
When the agent pulls arm $k_t$ at round $t$ and observes reward $r_t$, it
incurs a \emph{realized baseline regret}
$
\delta_t \;=\; \mu_0 - r_t,
$
measuring the shortfall of the pulled arm's reward relative to the baseline
mean. Because $\delta_t$ is defined against the realized reward $r_t$ rather than the arm mean $\mu_{k_t}$, it is a noisy, round-level quantity that can
be negative whenever $r_t$ exceeds $\mu_0$, at which point the pull \emph{replenishes} rather than depletes the budget. The cumulative realized regret after $n$ pulls is
\begin{equation}
\mathcal{R}_n \;=\; \sum_{t=1}^{n}\bigl(\mu_0 - r_t\bigr), \qquad
\mathcal{R}_0=0.
\end{equation}

\paragraph{Constraints.}
The agent operates under two simultaneous constraints. First, a \emph{regret budget} $B\ge 0$ bounds the cumulative realized regret at every pull along the trajectory:
\begin{equation}
\mathcal{R}_t \;\le\; B, \qquad \forall\, t\le\tau.
\end{equation}
Second, a \emph{pull budget} $T\in\mathbb{N}$ caps the total number of arm
pulls, so that $\tau\le T$. The pull budget is necessary for the problem to
be well posed: since $\delta_t$ can be negative, an agent that repeatedly
pulls a high-reward arm can drive $\mathcal{R}_n$ to be non-increasing, so
the regret constraint alone would permit indefinite sampling. The two
constraints together guarantee a genuine trade-off between information gained
and risk incurred, even when the regret budget does not bind.

\paragraph{Objective.}
The goal of $\pi$ is to maximize the probability of exactly identifying the
best arm, subject to both constraints:
\begin{equation}
\max_{\pi}\; \mathbb{P}\bigl(\phi = 1\bigr)
\;\text{s.t.}\quad
\mathcal{R}_t\le B,\ \forall\, t\le\tau,\quad \tau\le T.
\end{equation}

\section{Structural Properties of \gls{acr:sh}}

\gls{acr:sh} is a widely used elimination algorithm for fixed-budget \gls{acr:bai}. For simplicity, suppose throughout this section that $K$ is a power of $2$ and that there is a margin of safety in the budget to complete the last round-robin cycle. This standard simplification avoids the clutter associated with floors, ceilings, and incomplete round-robin cycles, without affecting the leading-order terms in the resulting bounds.  Starting from the full set of arms, \gls{acr:sh} proceeds over \(\log_2(K)\) phases. In each phase, the active arms are sampled in round-robin fashion, and the half with the largest empirical means is retained. Throughout,  denote by $n_q$ the number of arms in each phase~$q$, and, when \(n_q=2\), use the convention \(n_q/4+1=2\) and $\Delta_{2K}=1$.

This phase structure admits a natural extension to \gls{acr:bai} under resource constraints when the consumption associated with each pull is nonnegative. A pull quota and a resource quota are assigned to each phase. The active arms are then sampled until one of these quotas binds, after which the usual \gls{acr:sh} elimination step is performed. This is the principle underlying \gls{acr:shrr} introduced by \citet{li2024best}.

\paragraph{A conservative reduction to nonnegative consumption.}

Our regret-budgeted problem can be embedded into this generic framework. For each arm $i\in [K]$, we can define the positive-part regret consumption   $D_i^+$ as the random variable with distribution $(\mu_0-r_i)_{+}$, with $r_i\sim \nu_i$.
As the positive-part regret consumption dominates the regret consumption, any algorithm that respects a budget \(B\) for cumulative
positive-part regret also respects the original signed-regret constraint.
Applying \gls{acr:shrr} with \(D_i^+\) as the resource consumption therefore
provides an immediate feasible benchmark for our problem.

To state the resulting guarantee, let
$
d_i^{+}:=\mathbb{E}\!\left[D_i^+\right],
$
and let \(d_{(1)}^{+}\geq\cdots\geq d_{(K)}^{+}\) denote these expected
consumptions in nonincreasing order. Define
{\small
\[
f(d):=
\begin{cases}
\displaystyle \frac{2}{\log(1/d)}, & 0<d<e^{-2},\\[3mm]
e^2d, & e^{-2}\leq d\leq 1 {\text{ or }} d=0,
\end{cases}
\]
}
together with the complexity terms
\[
    H_{2,T}
    :=
    \max_{k>1}
    \frac{k}{\Delta_k^2}\text{, }
    H_{2,+}^{\mathrm{sto}}
    :=
    \max_{k>1}
    \frac{\sum_{j=1}^{k}f(d_{(j)}^{+})}{\Delta_k^2}.
\]
Finally, let
$
    \gamma_{+}
    :=
    \min\left\{
        \frac{T}{H_{2,T}},
        \frac{B}{H_{2,+}^{\mathrm{sto}}}
    \right\}.
$
Ignoring integer-rounding effects,
\gls{acr:shrr} satisfies
\begin{equation}
    \mathbb{P}(\phi\neq 1)
    \leq
    14K\log_2(K)\exp\left(-\frac{\gamma_{+}}{8\log_2(K)}\right).
    \label{eq:positive-part-benchmark}
\end{equation}

\begin{remark}In the Supplementary Material, we also analyze a phase-wise-estimate variant of \gls{acr:shrr}, in
which each elimination decision uses only the samples collected during the
current phase. This modification leaves the resource-rationing mechanism
unchanged and removes the extraneous factor \(K\) from the prefactor in
\eqref{eq:positive-part-benchmark}.
\end{remark}

\paragraph{What the positive-part reduction misses.}

Bound \eqref{eq:positive-part-benchmark} is a useful benchmark, but the
reduction on which it relies deliberately discards the distinctive structure
of regret. 

First, a pull whose reward exceeds the baseline has zero positive-part
consumption, whereas its signed consumption is negative. Such a pull
replenishes the original regret budget and can finance additional
exploration. This effect is entirely absent from the positive-part
formulation. For example, suppose that rewards for arm $i$ are Bernoulli and that
\(0<\mu_0<\mu_i<1\). Then
$ \mathbb{E}_{X_i\sim \nu_i}[\mu_0-X_i] =\mu_0-\mu_i <0,
$
whereas
$
    \mathbb{E}_{X_i\sim \nu_i}[(\mu_0-X_i)_+]
    =
    \mu_0(1-\mu_i)
    >0.
$
Thus, an arm that replenishes the signed-regret budget in expectation is
nevertheless assigned a strictly positive expected cost by the positive-part
reduction.

Second, generic resource-constrained analyses treat each arm's consumption
distribution as a fixed, exogenous attribute. Although the arm-level
distributions are also fixed in our setting, the consumption profile of the \emph{active set} changes endogenously as elimination removes poorly performing arms. In particular, the generic bound controls a phase containing \(n_q\) arms using the \(n_q\) largest effective consumptions
\(f(d_i^+)\), independently of which arms have actually survived. It
therefore charges each phase according to a worst-case consumption profile
of the corresponding cardinality.
Signed regret is more structured. Under round-robin sampling, the expected
regret consumption per pull in phase \(q\) is $\frac{1}{n_q}
    \sum_{i\in S_q}\Delta_i-\Delta_0$
where \(\Delta_0=\mu^\star-\mu_0\). Consequently, controlling the average
suboptimality of the surviving arms simultaneously controls the expected
resource consumption of the phase.
 
\paragraph{A regret-specific structural bound.}

We now analyze the signed-regret process directly. Consider a phase-rationed version of \gls{acr:sh} in which phase \(q\) is assigned a regret budget \(B^{(q)}\). The allocation \(B^{(q)}\) need not be uniform. The result below applies to any phase-budget allocation; the proportional allocation defining \gls{acr:shprr} is introduced in the next section.

For the theoretical analysis, elimination decisions are based on empirical means computed only from the observations collected during the current phase. Algorithm~\ref{alg:SHPRR}, which is used in our experiments, instead computes empirical means from all observations accumulated up to that phase. We analyze the phase-wise-estimate variant because it follows the standard theoretical formulation of \gls{acr:sh} and permits sharper guarantees, even though discarding past observations may degrade empirical performance. The two variants otherwise share the same sampling, stopping, and budget-rationing mechanisms.

Define, for each phase $q$:
\[
\underline{\tau_q}=\min\left(\frac{T}{n_q\log_2 K}; \frac{B^{(q)}}{6n_q\Delta_{2n_q}}\right),
\]
which we show is a conservative lower bound in the number of pulls per arm in phase $q$, 
and
\[
\Gamma_{2,R}=\min_{q\in [\log_2(K)]}\underline{\tau_q}\Delta_{\frac{n_q}{4}+1}^2.
\]

The central structural property is that the average gap of the surviving set
improves at a controlled rate.

\begin{proposition}[Phase-wise Arm Improvement]\label{prop:improvement}
  Suppose $\frac{\log 4e\log_2(K)^2}{\Delta_{\frac{n_q}{4}+1}^2}\leq\frac{1}{8}
   \underline{\tau_q} $. Then,  with probability at least:
    \[
    1-3\log_2(K)^2\exp\left(-\frac{\Gamma_{2,R}}{8\log_2(K)^2}\right),
    \]
   for every phase $q'$, we have:
\[
\frac{1}{n_{q'}}\sum_{i \in S_{q'}} \Delta_i \leq 3\Delta_{2n_{q'}}.
\]
\end{proposition}

This phase-wise improvement also yields a bound on the probability of
misidentification.
\begin{theorem}\label{thm:main}
Suppose $\frac{\log 4e\log_2(K)^2}{\Delta_{\frac{n_q}{4}+1}^2}\leq\frac{1}{8}
   \underline{\tau_q} $. Then:
\[
\mathbb{P}\left(\phi\neq 1\right)\leq 7\log_2(K)^2\exp\left(-\frac{\Gamma_{2,R}}{8\log_2(K)^2}\right)
\]
\end{theorem}
Unlike the positive-part benchmark, Theorem~\ref{thm:main} does not charge each phase according to a fixed worst-case consumption profile. Instead, the effective phase length adapts to the phase-specific quality scale $\Delta_{2n_q}$, which decreases as suboptimal arms are eliminated. Because regret can be negative and replenish the budget, standard lower-bound techniques do not directly apply; consequently, the theorem does not fully characterize the intrinsic complexity of pure exploration under a regret constraint. It nevertheless captures the declining cost of exploration across phases, while the algorithm further carries forward any budget accumulated through negative realized regret.

This observation motivates allocating a larger share of the regret budget to
the early, expensive phases. The next section introduces the proportional
rationing schedule used by \gls{acr:shprr}.

\section{\gls{acr:shprr} Algorithm}

In the regret-budgeted setting, an elimination algorithm's performance is largely determined by how it distributes its budget over time. Spending too cautiously at the start allows low-reward, high-consumption arms to survive into later phases, where each additional pull on a candidate that was never competitive drains budget needed for the discriminations that actually matter. Spending too aggressively might result in an insufficient reserve for separating the final, closely matched candidate arms. Proposition~\ref{prop:improvement} provides
the structural basis for resolving this tension: since the surviving set
improves from phase to phase, the per-pull cost of information declines over
the run, so the budget is most valuable, and should be concentrated, where
consumption is highest, namely in the early phases. At the same time, the appropriate degree of front-loading admits no universal answer: it depends on how quickly the surviving set improves and on how costly the initial pool is, both properties of the instance rather than the algorithm. The degree of front-loading should therefore itself be a design parameter, exposed to the practitioner rather than fixed in advance. 

To this end, we introduce \gls{acr:shprr}. Building on the \gls{acr:shrr} framework, \gls{acr:shprr} utilizes \textit{non-uniform rationing} of the regret budget, controlled by a tunable parameter $\alpha \in [0, 1]$. Unlike \gls{acr:shrr}, which would equally divide the regret budget across phases, \gls{acr:shprr} allocates it proportionally to the number of arms remaining in each phase, raised to the power~$\alpha$. With  $n_q = \lceil K / 2^{q-1} \rceil$, the regret budget allocated to phase $q$ is defined as:
\begin{align}
B_{q} = \mathcal{R}_q(K,B,\alpha) = B \cdot \frac{n_q^{\alpha}}{\sum_{j=1}^{\lceil \log_2 K \rceil} n_j^{\alpha}}. \label{ration}
\end{align}

\begin{algorithm}[t]
\caption{Sequential Halving with Proportional Regret Rationing (SHPRR)}
\label{alg:SHPRR}
\begin{algorithmic}[1]

\Require Regret budget $B\in\mathbb{R}_+$, total arm pull budget T, arm set $S$ of size K, baseline arm mean $\mu_{0}$.

\State $S^{(1)} \gets S$, $t \gets 1$. \label{SHPRR:1}
\State $B^{(1)} \gets \mathcal{R}_1(K,B,\alpha)$, $T^{(1)} \gets T/\lceil \log_2 K \rceil$. \label{SHPRR:2}
\For{$q=1$ \textbf{to} $\lceil \log_2 K \rceil$} \label{SHPRR:3}
  \State Initialize $I^{(q)}\gets 0$, $H^{(q)}\gets\emptyset$, $t^{(q)} \gets 1$. \label{SHPRR:4}
  \While{$I^{(q)} \le B^{(q)} - 1$ and $t^{(q)}\leq T^{(q)}-1$} \label{SHPRR:5}
    \State Pull arm $k_t$ based on round robin selection. \label{SHPRR:6}
    \State Observe reward $r_t$ and regret consumption~$\delta_t$. \label{SHPRR:7}
    \State Update $I^{(q)} \gets I^{(q)} + \delta_t$. \label{SHPRR:8}
    \State Update $H^{(q)} \gets H^{(q)} \cup \{(k_t,r_t)\}$.
    \State Update $t^{(q)} \gets t^{(q)}+1$, $t \gets t + 1$. \label{SHPRR:9}
  \EndWhile
  \State Use $\bigcup_{m=1}^{q} H^{m}$ to compute empirical means~$\{\hat{\mu}^q_{k}\}_{k \in S^{(q)}}$. \label{SHPRR:10}
  \State Let $S^{(q+1)}$ be the top $\lceil |S^{(q)}|/2 \rceil$ arms with highest empirical means. \label{SHPRR:11}
  \State Update \mbox{$B^{(q+1)} \gets \mathcal{R}_{q+1}(K,B,\alpha) + (B^{(q)} - I^{(q)})$.}\label{SHPRR:15}
  \State Update \mbox{$T^{(q+1)} \gets  T/\lceil \log_2 K \rceil + (T^{(q)} - t^{(q)})$.} \label{SHPRR:16}
\EndFor

\State \textbf{Return} the arm with the highest empirical mean in $S^{(q)}$. \label{SHPRR:17}

\end{algorithmic}
\end{algorithm}

When $\alpha = 0$, this recovers uniform allocation across phases, as in \gls{acr:shrr}. As $\alpha$ approaches 1, more resources are allocated to earlier phases, in proportion to the number of arms considered in each phase. For $0<\alpha<1$, the allocation decays exponentially across phases but at a shallower rate than at $\alpha=1$, giving moderate preference to early phases while retaining more resources for later ones than full proportional ($\alpha=1$) allocation would.

Algorithm~\ref{alg:SHPRR} presents the \gls{acr:shprr} procedure for identifying the optimal arm under regret-based and maximum arm pull resource constraints. The algorithm proceeds over $\lceil \log_2 K \rceil$ phases, where $K$ denotes the initial number of arms. The active arm set $S^{(1)}$ and the time step counter are first initialized (Line~\ref{SHPRR:1}). In Line~\ref{SHPRR:2}, the total regret budget $B$ is proportionally allocated across all phases according to Expression~\ref{ration}, while the total arm pull budget $T$ is split uniformly across phases, giving each phase an initial pull allowance $T/\lceil \log_2 K \rceil$. Within each phase $q$ (Lines~\ref{SHPRR:4}--\ref{SHPRR:9}), \gls{acr:shprr} samples arms from the active set $S^{(q)}$ in round-robin order, continuing until \emph{either} the phase's regret budget $B^{(q)}$ or its arm pull budget $T^{(q)}$ is exhausted, whichever binds first. Each pull's regret consumption is measured as the deviation of the observed reward from the baseline mean $\mu_0$ (Line~\ref{SHPRR:7}). This quantity is negative whenever an arm outperforms the baseline, in which case the pull replenishes rather than depletes the phase's regret budget. Once sampling stops, empirical means are computed using the full pull history accumulated across all phases so far (Line~\ref{SHPRR:10}), and the top half of the active arms by empirical mean advance to the next phase (Line~\ref{SHPRR:11}), following the standard \gls{acr:sh} elimination rule.

Before the next phase begins, both budgets are updated (Lines~\ref{SHPRR:15}--\ref{SHPRR:16}). The regret budget for phase $q+1$ combines a fresh allocation from Expression~\ref{ration}, sized to the new number of active arms, with any unspent or replenished regret budget carried over from phase $q$; the arm pull budget is updated analogously, combining a fresh uniform allocation with any unused pulls from the previous phase. This carry-over mechanism ensures that budget conserved in early phases, whether through early elimination of costly arms or through high-reward arms replenishing the budget, remains available for the more difficult discriminations of later phases, rather than being lost. The procedure terminates after $\lceil \log_2 K \rceil$ phases and returns the arm with the highest empirical mean among the final surviving arms (Line~\ref{SHPRR:17}).

\section{Experiments}
In this section, we present our experimental design, followed by a performance overview of \gls{acr:shprr} against benchmark algorithms, a sensitivity analysis, and an examination of how varying the rationing parameter affects the arm composition retained at each phase.

\subsection{Experimental Design}
We evaluate our method on synthetic and real-world benchmark datasets to demonstrate its effectiveness across a range of reward structures, gap configurations, and consumption profiles. Following~\citet{li2024best}, our synthetic environments use four mean-reward configurations of Bernoulli arms (OneSub, Trap, Poly, and Geom). Additionally, we consider variations of these environments where we vary the optimal--suboptimal gap via $\mu_k' = \max(\mu_k - \delta, 0)$ for all $k \neq k^\star$, with $\delta \in \{0, 0.1, 0.2, 0.3\}$, where $\delta=0$ recovers~\citet{li2024best}. We define the regret consumption of each arm relative to the top 3\% quantile arm, and we use a regret budget of 384 for the performance overview and 224 for the per-phase analysis, a max pull budget of 50000, and 500 seeds per configuration.

We complement these with two real-world datasets reflecting human preferences. In the New Yorker Caption Contest \cite{jain2020newyorker}, arms are submitted captions with binary rewards reflecting whether voters found them funny, with mean reward given by the empirical positive-vote rate; we select three contests (511, 538, and 780, denoted c511, c538, and c780) with 3,000--6,000 arms each, chosen for their distinct reward distributions to cover a range of real-world cases. We use the top~1\% quantile arm as baseline, a regret budget of 1.6 times the number of arms in each contest, a max pull budget of 100,000, and 500 seeds. In the Criteo Attribution Modeling for Bidding dataset \cite{DiemertMeynet2017}, arms are ad campaigns; pulling campaign $i$, with estimated attribution probability $\hat{p}_i$ and mean conversion value $v_i$, yields $A_t^{(i)} \sim \mathrm{Bernoulli}(\hat{p}_i)$ and reward $r_t^{(i)} = v_i A_t^{(i)}$. We select this dataset for its very low expected rewards (maximum arm mean 0.015), a challenging low-signal setting. Consumption is again the regret gap to a quantile baseline, tested at the top~5\%, top~1\%, and best-arm baselines; and we sweep the regret budget from 0.06 to 1.8 times the number of arms. All Criteo experiments use a max pull budget of 100,000 and 500 random seeds, with further design details in the Supplementary Material.

We compare \gls{acr:shprr} against three state-of-the-art algorithms: \gls{acr:shrr}, \gls{acr:tt}, and AT-LUCB. Since \gls{acr:tt} and AT-LUCB were not originally designed to respect a resource budget, we adapt both by adding budget exhaustion as an additional stopping criterion. Note that, even though \gls{acr:shrr} originally assumes strictly positive consumption, we evaluate all three benchmarks under our setting: each operates under the same signed regret consumption $\delta_t = \mu_0 - r_t$ and thus benefits equally from budget replenishment when arms outperform the baseline. We report the average \gls{acr:sp}, defined as the fraction of runs identifying the optimal arm, and the average \gls{acr:nsr}, defined as $\frac{\mu^* - \mu^{j}}{\mu^* - \mu^{\min}}$, where $\mu^*$, $\mu^j$, and $\mu^{\min}$ denote the optimal, selected, and worst arm means respectively.
\begin{figure}[t!] 
\centering 
\includegraphics[width=0.95\linewidth]{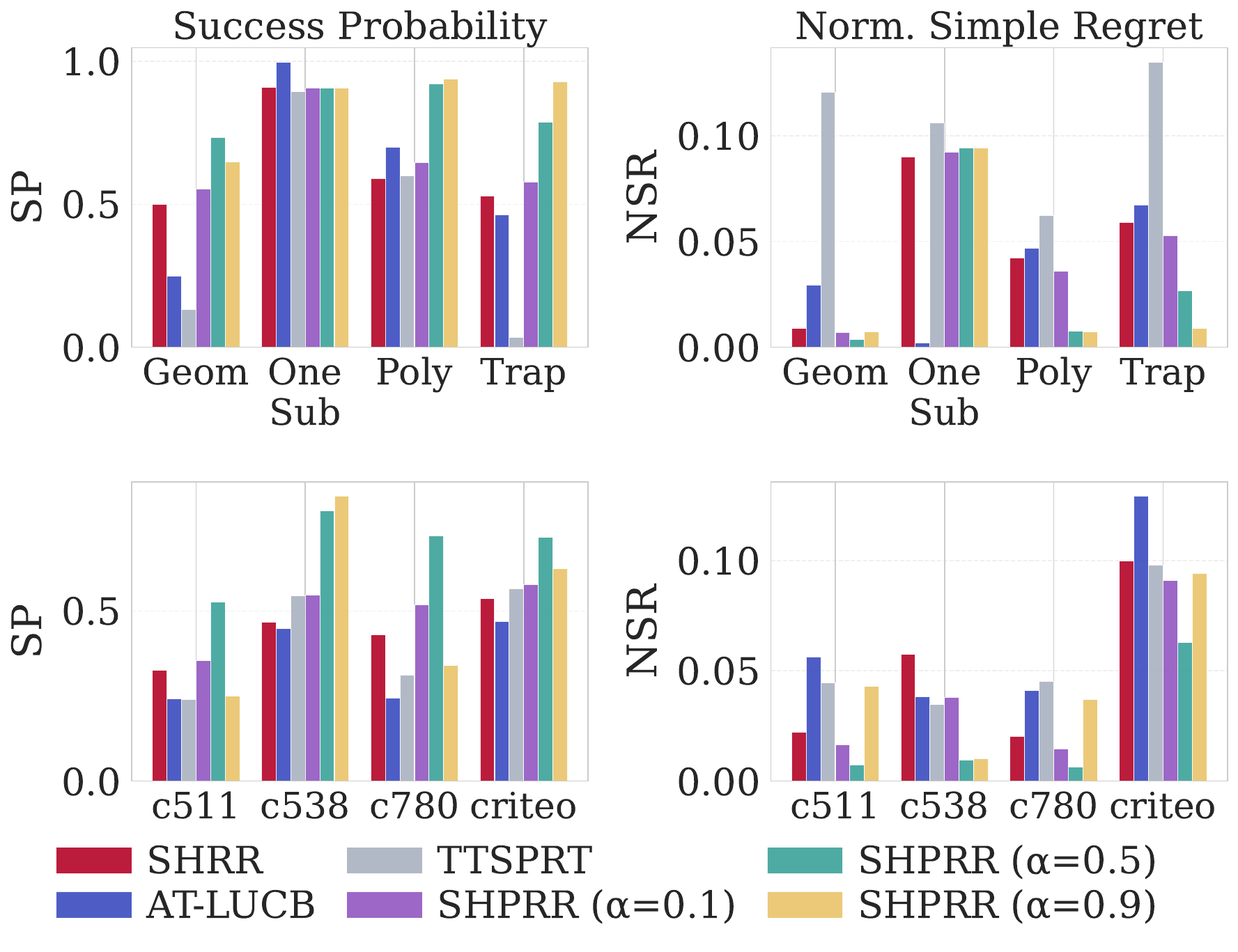}\caption{\gls{acr:sp} and \gls{acr:nsr} across synthetic (\textit{top}) and real-world (\textit{bottom}) environments.}\label{fig:perf_overview} \end{figure}

\subsection{Performance Overview}

Figure~\ref{fig:perf_overview} reports \gls{acr:sp} and \gls{acr:nsr} for all methods across the four synthetic environments (top) and the real-world datasets (bottom), with the baseline arm set at the top 1\% quantile. Overall, \gls{acr:shprr} frequently outperforms the state-of-the-art benchmarks across all four synthetic environments. \gls{acr:shprr} with higher rationing ($\alpha = 0.5, 0.9$) achieves the best or near-best performance in Geom, Poly, and Trap, yielding both higher \gls{acr:sp} and lower \gls{acr:nsr} than all benchmarks. In Trap, \gls{acr:shprr} ($\alpha = 0.9$) attains an \gls{acr:sp} of 0.93, nearly double that of \gls{acr:shrr} (0.528) and AT-LUCB (0.462), while \gls{acr:tt} collapses to 0.034. A similar trend holds in Geom and Poly, where $\alpha>0$ improves both \gls{acr:sp} and \gls{acr:nsr}. The sole exception is OneSub, a degenerate single-dominant-arm case where AT-LUCB attains near-perfect identification (\gls{acr:sp} 0.998, \gls{acr:nsr} 0.002) and all \gls{acr:shprr} variants and \gls{acr:tt} perform comparably (\gls{acr:sp} 0.89--0.91); here the consumption gap is homogeneous across suboptimal arms by construction, offering no heterogeneity for rationing to exploit. Overall, \gls{acr:shprr} is best or near-best in every environment with non-trivial consumption structure and maintains consistently strong performance across all four environments, whereas the baseline algorithms fluctuate considerably: AT-LUCB and \gls{acr:tt} each achieve near-best results in one environment (OneSub) but degrade sharply in others (e.g., AT-LUCB falls to 0.248 in Geom, and \gls{acr:tt} falls to 0.034 in Trap), underscoring the robustness of our approach across varying consumption structures.
\begin{figure}[t!]
\centering
\includegraphics[width=0.95\linewidth]{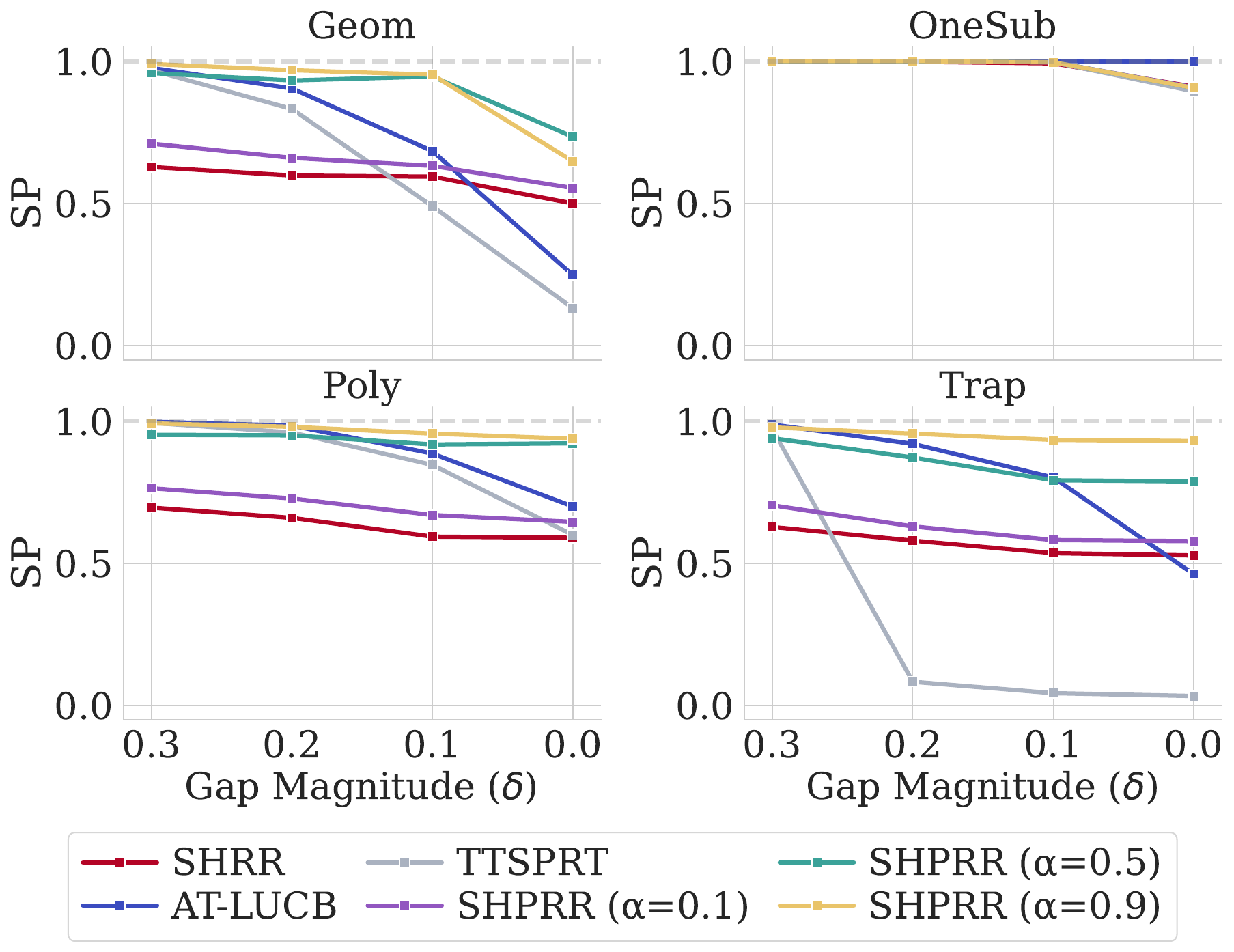}
\caption{\gls{acr:sp} across the four synthetic environments as $\delta$ varies from 0.3 to 0. \label{fig:sensitivity_delta}}
\end{figure}
On the real-world datasets \gls{acr:shprr} continues to outperform all benchmarks, though the best-performing rationing level varies more than in the synthetic environments. \gls{acr:shprr} ($\alpha=0.5$) achieves the highest \gls{acr:sp} in three of the four datasets, reaching 0.526 on caption contest 511, 0.72 on 780, and 0.716 on criteo, consistently outperforming all benchmarks and substantially reducing \gls{acr:nsr}. On caption contest 538, \gls{acr:shprr} ($\alpha=0.9$) performs best among all methods; however, front-loading too aggressively can result in decreased performance, as observed in contests 511 and 780.

Importantly, although the optimal $\alpha$ varies by environment, a single fixed choice of $\alpha=0.5$ already strongly outperforms every benchmark in all but one of the synthetic and real-world environments. The tunability of $\alpha$ is therefore not a burden: moderate rationing ($\alpha=0.5$) serves as a strong, robust default when the reward structure is unknown, while the flexibility of $\alpha$ lets practitioners extract further gains when domain knowledge is available.

\subsection{Sensitivity Analysis}
\begin{figure}[t]
\centering
\includegraphics[width=\linewidth]{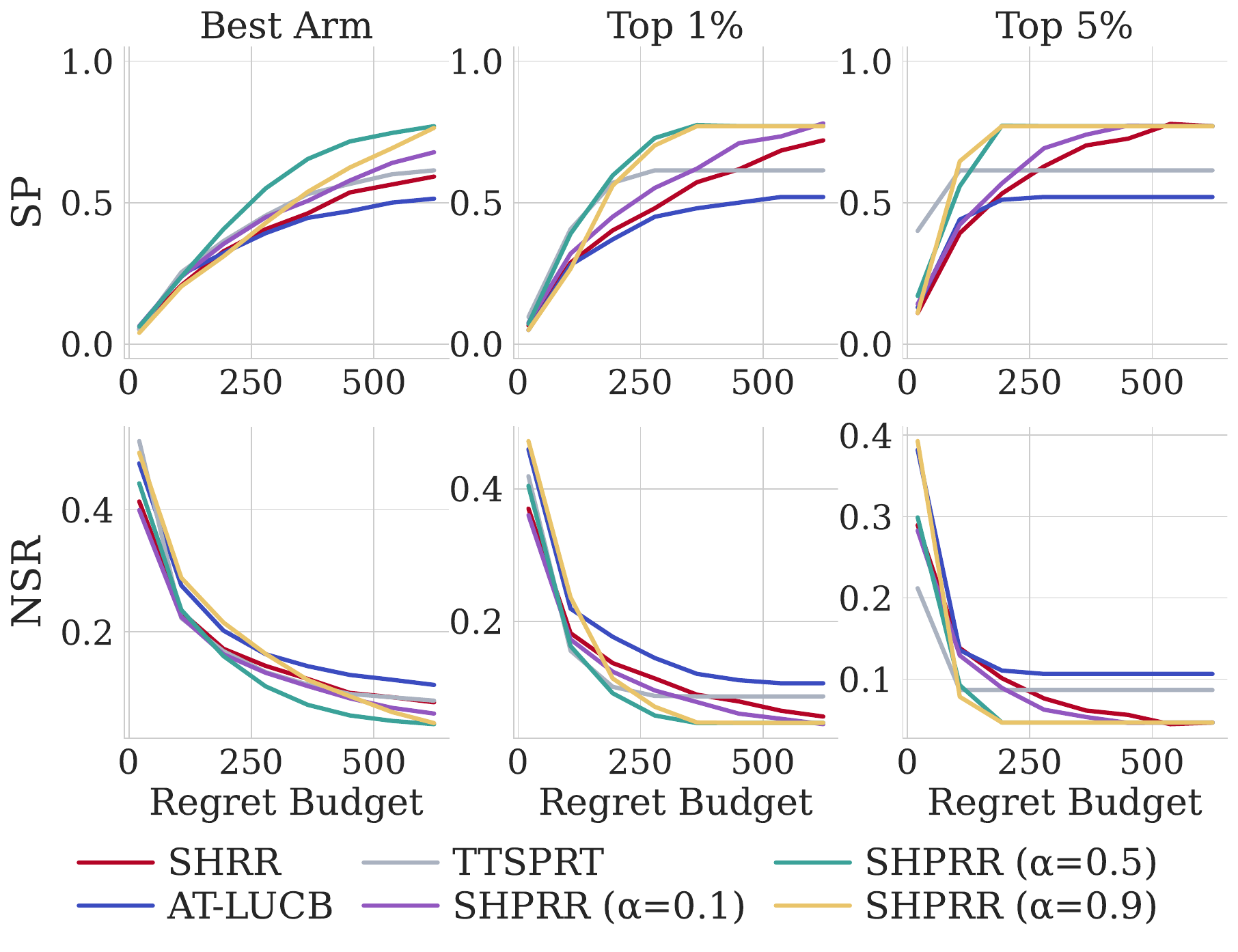}
\caption{Performance analysis across three baseline arm selection criteria (from left to right): best arm, top~1\%, and top~5\%, evaluated under varying budget levels. \label{fig:baseline_arm}}
\vspace{-0.2cm}
\end{figure}

\subsubsection{Gap Magnitude.}
An algorithm intended for real-world deployment cannot rely on favorable conditions and must maintain stable performance across diverse instances. We therefore examine sensitivity to the optimal--suboptimal gap $\delta$, which controls reward similarity as $\delta \to 0$. Figure~\ref{fig:sensitivity_delta} shows a consistent pattern: while the baselines are competitive only when the gap is large and the problem is easy, \gls{acr:shprr} pulls decisively ahead as $\delta \to 0$ where identification becomes harder, degrading far more gracefully than every benchmark. This is clearest in Trap, where \gls{acr:shprr} ($\alpha = 0.9$) stays above $0.93$ across the entire range, while AT-LUCB falls from $0.988$ to $0.462$ and \gls{acr:tt} collapses from $0.972$ to $0.034$. The same pattern holds in Geom, where the best \gls{acr:shprr} variant remains at $0.734$ at $\delta = 0$ while AT-LUCB falls to $0.248$ and \gls{acr:tt} to $0.132$, and in Poly, where \gls{acr:shprr} ($\alpha = 0.9$) holds $0.938$ against AT-LUCB's $0.70$ and \gls{acr:tt}'s $0.60$. OneSub is the sole exception: with a single dominant arm, all methods remain near ceiling regardless of $\delta$. Overall, these results indicate that \gls{acr:shprr} is the only dependable choice: the single method that maintains strong performance independent of the optimal--suboptimal gap.

\subsubsection{Baseline Arm and Regret Budget.}
We further examine how performance varies with the choice of baseline arm that defines consumption, comparing best-arm, top~1\%, and top~5\% baselines on the Criteo dataset across a range of regret budgets (Figure~\ref{fig:baseline_arm}). The best-arm case is the most demanding: every suboptimal arm incurs positive consumption and none can replenish the budget, recovering the strictly non-negative, non-replenishing structure of the original \gls{acr:bai}-under-constraints formulation. Even in this subcase, \gls{acr:shprr} ($\alpha=0.5$) outperforms all benchmarks steadily across budget levels, reaching the highest \gls{acr:sp} (0.77) and lowest \gls{acr:nsr} at the largest budget, while AT-LUCB trails throughout. As the baseline is relaxed to the top~1\% and top~5\% quantiles, an increasing number of arms outperform the baseline and thus replenish rather than consume the budget. Here \gls{acr:shprr} ($\alpha=0.5, 0.9$) saturates at the highest \gls{acr:sp} (0.77) and uses a smaller fraction of the budget than \gls{acr:tt} and AT-LUCB, which plateau early. This shows that \gls{acr:shprr}'s rationing more efficiently exploits budget-replenishing arms, converging to a high-quality solution well before the pull budget is exhausted. Across all three baselines, \gls{acr:shprr} is the only method that remains strong whether the budget is scarce and non-replenishing or lenient and replenishable.

\subsection{Per Phase Arm Composition}
\begin{figure}[t!]
\centering
\includegraphics[width=0.5\textwidth]{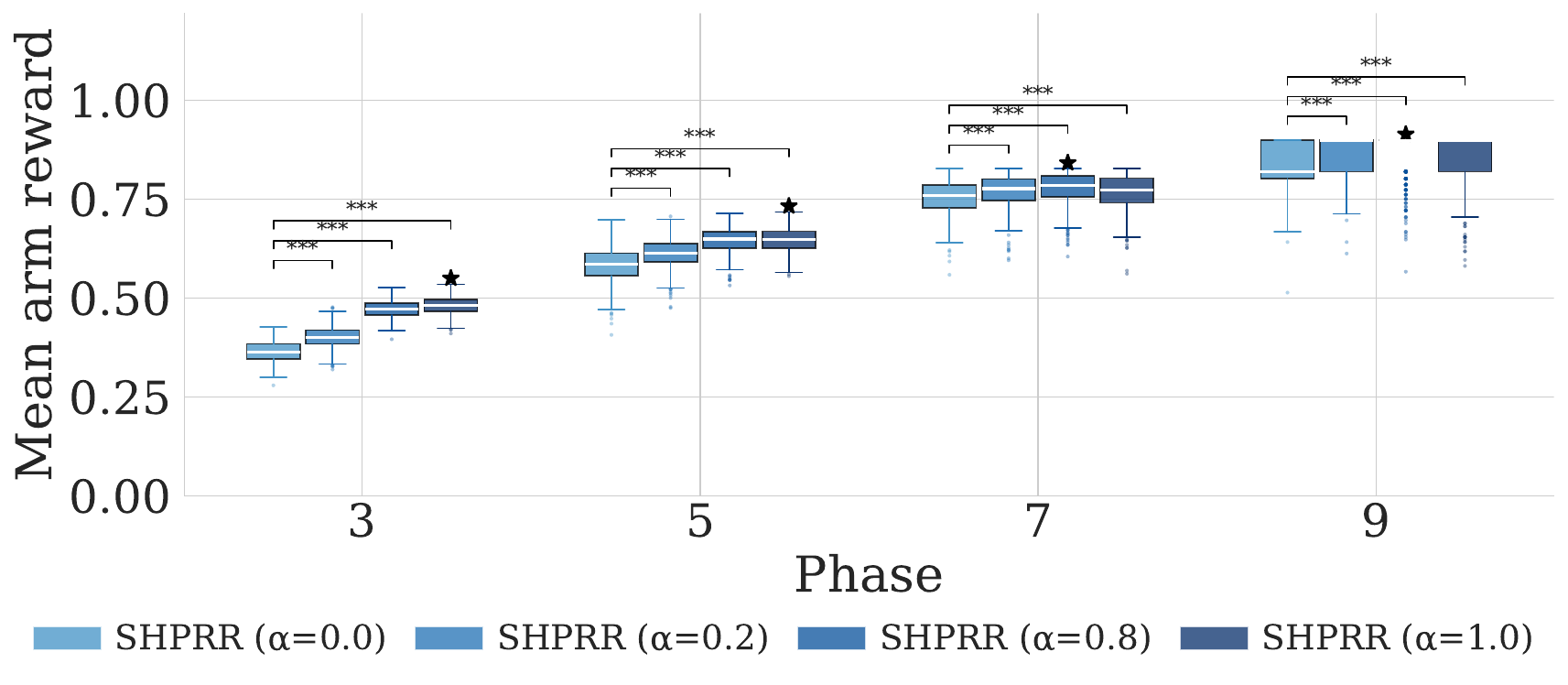}
\caption{Phase-wise arm quality under a Poly reward configuration. Boxplots show the mean reward distribution among surviving arms at each phase, for four \gls{acr:shprr} variants with $\alpha \in \{0.0, 0.2, 0.8, 1.0\}$. Significance brackets indicate $t$-tests paired across seeds against the $\alpha=0.0$ reference (\gls{acr:shrr}); a star ($\star$) marks the highest mean at each phase. Phase~9 is the final recommended arm.
\label{fig:poly_variant_0p5}}
\vspace{-0.4cm}
\end{figure}
To illustrate the effect of rationing, we track the mean reward of arms still active at each phase under the Poly reward configuration, comparing four \gls{acr:shprr} variants spanning the full rationing spectrum, from $\alpha = 0$ (\gls{acr:shrr}) to $\alpha = 1$ (maximal front-loading). Since arms are progressively eliminated, the mean reward of surviving arms rises over time regardless of~$\alpha$; the question is whether higher rationing accelerates this concentration toward high-reward arms, and
does so consistently across phases.

Figure~\ref{fig:poly_variant_0p5} shows that this is indeed the case: at every displayed phase, increasing $\alpha$ significantly raises the mean reward of surviving arms relative to the $\alpha=0.0$ reference. The effect is most pronounced in early phases, indicating that rationing helps the algorithm discard low-reward arms earlier in the run, but the effect is maintained throughout. By the end of the run (phase 9), the gap narrows somewhat as all variants converge toward high-reward arms, though $\alpha=0.8$ still attains the highest mean reward among the variants. These results reveal the mechanism behind \gls{acr:shprr}'s empirical gains: rationing actively steers early elimination toward high-reward arms and sustains that advantage throughout the run. Since consumption is the reward recentered against the baseline, concentrating the surviving set on high-reward arms directly lowers per-pull consumption in later phases, so the very mechanism that improves identification also relieves the budget, grounding the higher \gls{acr:sp} and lower \gls{acr:nsr} of moderate-to-high $\alpha$ in a concrete, measurable effect.

\section{Conclusion}
We introduced \gls{acr:bai} under a realized regret budget, where experimentation cost is measured as regret relative to a trusted baseline, yielding signed, reward-coupled, replenishable consumption that lies outside existing formulations. We proved that the per-pull price of information declines across the phases of \gls{acr:sh}, and exploited this property to design \gls{acr:shprr}, which front-loads the budget via a tunable parameter $\alpha$. It is the only method robust across all synthetic and real-world environments, with $\alpha=0.5$ a strong default.

\clearpage
\bibliography{aaai2027}

\end{document}


\maketitle

\section{Sharpening the results of the paper \cite{li2024best}}

Classical Sequential Halving bases each elimination decision on the observations collected during the current phase. In contrast, SH-RR as defined by \cite{li2024best} uses cumulative empirical means. We analyze the phase-wise-estimate variant of SH-RR, obtained all empirical means with empirical means computed from the current-phase history
only. This modification leaves the resource-rationing mechanism
unchanged and permits the standard expected-count and Markov argument used in the analysis of Sequential Halving. This allows us to sharpen the dependency in the number of arms in the prefactor of each bounds, specifically removing an extraneous factor $K$. Although this version of the algorithm does not necessarily work better in practice, it eases the comparison of the complexity of the problem with other variants seen in the literature.

In the stochastic-consumption setting, we complement this argument with an aggregate concentration inequality for the total resource consumption in a phase, which avoids the additional armwise union bound in the original proof. 

We state the resulting guarantees in the general BAIwRC model with $L$ resource constraints, for completeness.

\textbf{Setting}: There is $L$ ressources, each with a budget $C_\ell$. A pull of arm $i$ generates a consumption $D_{\ell, i}$ of resource $\ell$. In the deterministic case, $D_{\ell, i}= d_{\ell,i}$ deterministically. In the stochastic case, $D_{\ell, i}\in [0;1]$ a.s, while $\mathbb{E}[D_{\ell, i}]=d_{\ell,i}$. The rest is the same as in the main text. Note that the original paper assumes that the arm rewards are $1$-subgaussian, while we assume that the rewards lie in $[0,1]$. The stronger assumption is necessary when one of the resources can be the regret. To avoid confusion, we keep it, though the same bound up to a modified concentration constant could be obtained with $1$-subgaussian rewards if none of the ressources are the regret.

\textbf{Notations}: Throughout the Appendix, we denote $t_q$ the (possibly random) number of pulls per arm in phase $q$. We define $\hat{\mu}_i^q$ as the empirical mean for arm $i$ at the end of phase $q$.

\subsection{The deterministic consumption bound}

Define denote $d_{\ell,i}$ the permutation of the indices so that $d_{\ell,(1)}\geq , \ldots, \geq d_{\ell,(K)}$.
\[
H_{2,\ell}:= \max_{k\in [K]} \frac{\sum_{j=1}^kd_{\ell,(j)}}{\Delta_k^2}, 
\]

and 
\[
\gamma(Q) = \min_{\ell \in [L]}\frac{C_\ell}{H_{2,\ell}}.
\]

We prove the following theorem, which sharpens the pre-factors' dependency in the number of arms in the bound in \cite{li2024best} from $K\log_2(K)$ to $\log_2(K)$.
\begin{theorem}\label{thm:improvedthmdeterministic}
Under the deterministic consumption setting,
\[
\mathbb{P}\left(\phi \neq 1\right)\leq 3\log(K)  \exp\left(-\frac{\gamma(Q)}{8\log{K}}\right)
\]
\end{theorem}

\begin{proof}
Note that in the deterministic consumption case, conditioned on $\mathcal{F}_q$, $t_q$ is not a random quantity. Therefore, by applying Hoeffding's inequality, we obtain the following lemma.
\begin{lemma}\label{lem:Hoeffding}
Assume that the best arm was not eliminated prior to round $q$. Then for any arm $i \in S^{(q)}$,
$$
\operatorname{Pr}\left[\hat{\mu}_1^q<\hat{\mu}_i^q|\mathcal{F}_q\right] \leq \exp \left(-\frac{1}{2} t_q \Delta_i^2\right) .
$$
\end{lemma}

We next bound the probability that the algorithm makes a mistake and excludes the best arm on phase $q$.
\begin{lemma}\label{lem:eliminiationroundr}
The probability that the best arm is eliminated on phase $q$ is at most
$$
3 \exp \left(-\frac{1}{8 \log _2 K} \cdot \gamma(Q)\right).
$$
\end{lemma}
\begin{proof} Suppose first that $|S^{(q)}|\geq 4$. We treat $|S^{(q)}|= 2$ separately.
 Define $S^{\prime(q)}$ as the set of arms in $S^{(q)}$, excluding the $\frac{1}{4}\left|S^{(q)}\right|=K / 2^{q+2}$ arms with the largest mean. Denote:
 \[
 i_q= \frac{1}{4}\left|S^{(q)}\right|+1.
 \]
 If the best arm is eliminated in phase $q$, it must be the case that at least half the arms of $S^{(q)}$ (i.e., $\frac{1}{2}\left|S^{(q)}\right|=n / 2^{q+1}$ arms) have their empirical average larger than its empirical average. In particular, the empirical means of at least $\frac{1}{3}\left|S^{\prime(q)}\right|=n / 2^{q+2}$ of the arms in $S^{\prime(q)}$ must be larger than that of the best arm at the end of phase $q$. Letting $N_q$ denote the number of arms in $S^{\prime(q)}$ whose empirical average is larger than that of the optimal arm, we have by Lemma \ref{lem:Hoeffding}:

$$
\begin{aligned}
\mathbf{E}\left[N_q\right] & =\sum_{i \in S^{\prime(q)}} \operatorname{Pr}\left[\hat{\mu}_1^q<\hat{\mu}_i^q\right] \leq \sum_{i \in S^{\prime(q)}} \exp \left(-\frac{1}{2} t_q \Delta_i^2\right) \\
& \leq   |S^{\prime(q)}| \exp \left(-\frac{1}{2} t_q \Delta_{i_q}^2\right)
\end{aligned}
$$
We now lower bound $t_q$. By design of the algorithm:
\begin{align*}
t_q \geq &\min_{\ell}  \frac{C_{\ell}}{\log_2(K)} \frac{1}{\sum_{j\in S^{(q)}} d_{\ell,j} }\\
\geq &\min_{\ell}  \frac{C_{\ell}}{\log_2(K)} \frac{1}{\sum_{j=1}^{|S^{(q)}|} d_{\ell,(j)} }\\
\geq &\min_{\ell}  \frac{C_{\ell}}{\log_2(K)} \frac{1}{4\sum_{j=1}^{i_q-1} d_{\ell,(j)} }.
\end{align*}
Plugging into the previous equation we obtain:
$$
\begin{aligned}
\mathbf{E}\left[N_q\right] 
& \leq   |S^{\prime(q)}| \exp \left(-  \frac{1}{8\log_2(K)} \min_\ell\frac{C_\ell}{\frac{1}{\Delta_{i_q}^2}\sum_{j=1}^{i_q-1}d_{\ell,(j)}}\right)\\
& \leq   |S^{\prime(q)}| \exp \left(-  \frac{1}{8\log_2(K)} \min_\ell\frac{C_\ell}{H_{2,\ell}}\right)\\
& \leq   |S^{\prime(q)}| \exp \left(-  \frac{1}{8\log_2(K)} \gamma(Q)\right),
\end{aligned}
$$
where the second inequality holds by definition of $H_{2,\ell}$ and the third by definition of $\gamma(Q)$.

We now apply Markov's inequality to obtain 
$$
\begin{aligned}
\operatorname{Pr}\left[N_q>\frac{1}{3}\left|S^{\prime(q)}\right|\right] & \leq 3 \mathbf{E}\left[N_q\right] /\left|S^{\prime(q)}\right| \\
& \leq 3 \exp \left(-\frac{1}{8 \log _2 K} \cdot \gamma(Q)\right)
\end{aligned}
$$

and the lemma is established for all $q$ such that $n_q\geq 4$. It remains only the case $n_q=2$. In that case, provided arm $1 \in S^{(q)}$, with $i$ the other arm in $S^{(q)}$:
\[
\mathbb{P}\left(\phi\neq 1|\mathcal{F}_q\right)=
\operatorname{Pr}\left[\hat{\mu}_1^q<\hat{\mu}_i^q|\mathcal{F}_q\right] \leq \exp \left(-\frac{1}{2} t_q \Delta_i^2\right)\leq  \exp\left(-\frac{1}{2} t_q \Delta_2^2\right).
\]
Also:
\begin{align*}
t_q \geq &\min_{\ell}  \frac{C_{\ell}}{\log_2(K)} \frac{1}{\sum_{j=1}^{2} d_{\ell,(j)}  }.
\end{align*}
Hence:
\[
\mathbb{P}\left(\phi\neq 1|\mathcal{F}_q\right)\leq \exp\left(-\frac{1}{2\log_2(K)} \min_{\ell}  \frac{C_\ell}{\frac{1}{\Delta_2^2}\sum_{j=1}^{2} d_{\ell,(j)} }\right)\leq 3\exp\left(-\frac{1}{8\log_2(K)}\gamma(Q)\right).
\]
\end{proof}
The proof of Theorem \ref{thm:improvedthmdeterministic} is now immediate.

Proof. Clearly, the algorithm does not exceed the budget of any resource. Also, if the best arm survives the execution, then the algorithm succeeds as all other arms must be eliminated after $\log _2 n$ phases. Finally, by Lemma \ref{lem:eliminiationroundr} and a union bound, the best arm is eliminated in one of the $\log _2 K$ phases of the algorithm with probability at most $3 \log _2 K \cdot \exp \left(-\frac{\gamma(Q)}{8  \log _2 K}\right)$,
which gives the theorem.
\end{proof}

\subsection{Stochastic consumption setting}

\begin{theorem}
    Consider a BAIwRC instance $Q$ in the stochastic consumption setting. The SH-RR algorithm has BAI failure probability $\operatorname{Pr}(\phi \neq 1)$ at most

$$
(L+3) \left(\log _2 K\right) \exp \left(-\frac{1}{8\left\lceil\log _2 K\right\rceil} \cdot \gamma^{\text {sto }}(Q)\right),
$$

where $\gamma^{\text {sto }}(Q)=\min _{\ell \in[L]}\left\{C_{\ell} / H_{2, \ell}^{\text {sto }}(Q)\right\}$ and $$
H_{2, \ell}^{\mathrm{sto}}(Q)=\max _{k \in\{2, \ldots, K\}}\left\{\frac{\sum_{j=1}^k f\left(d_{\ell,(j)}\right)}{\Delta_k^2}\right\}.
$$

\end{theorem}
\begin{proof}

For the stochastic setting, the main difficulty is that the number of pulls is stochastic. The proof then needs to lower bound the total number of pulls and also take into account that the number of samples in the empirical mean are history dependent.

We have two helper lemmas. The first one will be helpful to lower bound the number of pulls in each phase. It is a modified version of Lemma 7 in \cite{li2024best}. The lemma in the original paper helps control the consumption per arm, this modified version controls the aggregate consumption. Again, the goal is to avoid resorting to a union bound.

\begin{lemma}\label{lem:lowerboundnumsamples1}
    Consider a family $(X_{i,s})_{i \in [m],s\in [N]}$ of mutually independent rewards in $(0,1]$ such that for each $i$, $\mathbb E [X_{i,s}]= d_i$. 
Then
\[
\mathbb{P}\left(
    \sum_{s=1}^{N}\sum_{i=1}^{m}X_{i,s}
    \geq N\sum_{i=1}^{m}f(d_i)
\right)
\leq e^{-N/2}.
\]
\end{lemma}

The function $f$ is chosen so that:
\begin{claim}\label{claim:LBKL}
    For any $d\in (0;e^{-2}]$, $\kl(f(d),d)\geq \frac{1}{2}$, where $\kl$ denotes the Bernoulli relative entropy.
\end{claim}
This claim, which will be useful for the proof of the lemma, is proved in the paper \cite{li2024best}, and we omit it here.

We can now show the number of samples in each phase is lower bounded.
    For each stage $q$, define 
\[
\underline{\tau_q} = \min_\ell \frac{C_\ell}{\log_2(K)\sum_{j=1}^{|S^{(q)}|}f\left(d_{\ell,(j)}\right)}.
\]
\begin{lemma}\label{lem:LBnbsamples}
For every phase $q$, conditionally on the history $\mathcal F_q$ at the
start of the phase,
\[
\mathbb P\left(t_q<\underline\tau_q\mid\mathcal F_q\right)
\leq L\exp\left(-\frac{\underline\tau_q}{2}\right).
\]
\end{lemma}

The second ingredient to the proof is a version of Lemma \ref{lem:Hoeffding} that holds when the number of samples is adaptive. 
\begin{lemma}\label{lem:adaptiveHoeffding}
    Let $1 \leq \tau \leq T$, and let $(X_i)_{i=1}^\infty$ and
    $(Y_i)_{i=1}^\infty$ be two mutually independent sequences of
    independent and identically distributed random variables taking values
    in $[0,1]$. Suppose that
    \[
        \Delta = \mathbb{E}[X_1-Y_1] > 0.
    \]
    Then
    \[
        \mathbb{P}\left(
            \exists t \in \{\tau,\ldots,T\} :
            \overline{X}_t \leq \overline{Y}_t
        \right)
        \leq \exp\left(-\tau\Delta^2\right).
    \]
\end{lemma}

With these two lemmas, we are ready to prove the main theorem. 

Let $\mathcal F_q$ denote the history at the beginning of phase $q$, and denote on $\mathcal E_q$ the event $\{1\in S^{(q)}\}$ that
the best arm has survived so far. 

First suppose that $n_q\geq 4$.  Put
\[
i_q:=\frac{n_q}{4}+1,
\]
and  define $S^{\prime(q)}$ as the set of arms in $S^{(q)}$, excluding the $\frac{1}{4}n_q=K / 2^{q+2}$ arms with the largest mean.
Every arm $i\in S'^{(q)}$ has global reward rank at least
$i_q$, and hence
\[
\Delta_i\geq \Delta_{i_q}.
\]
Let
\[
N_q:=\sum_{i\in S'^{(q)}}
\mathbf 1\!\left\{\widehat\mu_i^q\geq\widehat\mu_1^q\right\}.
\]
 If the best arm is eliminated in phase $q$, it must be the case that at least half the arms of $S^{(q)}$ have their empirical average larger than its empirical average. In particular, this implies
\[
N_q\geq \frac{n_q}{4}=\frac{|S'^{(q)}|}{3}.
\]
Therefore,
\begin{align*}
\mathbb P\left(\overline{\mathcal{E}}_{q+1}
\,\middle|\,\mathcal F_q, \mathcal{E}_q\right)\leq&
\mathbb P(t_q<\underline{\tau_q}\mid\mathcal F_q, \mathcal{E}_q)
+
\mathbb P\left(
N_q\geq \frac{|S'^{(q)}|}{3},\ t_q\geq\underline{\tau_q}
\,\middle|\,\mathcal F_q, \mathcal{E}_q
\right)
\end{align*}
By Lemma \ref{lem:LBnbsamples},
\[
\mathbb P(t_q<\underline{\tau_q}\mid\mathcal F_q, \mathcal{E}_q)\leq L\exp\left(-\frac{\underline{\tau_q}}{2}\right).
\]
By Markov's inequality:

\begin{align*}
\mathbb P\left(
N_q\geq \frac{|S'^{(q)}|}{3},\ t_q\geq\underline{\tau_q}
\,\middle|\,\mathcal F_q, \mathcal{E}_q
\right)\leq&
\frac{3}{|S'^{(q)}|}
\mathbb E\left[N_q\mathbf 1\{t_q\geq\underline{\tau_q}\}
\,\middle|\,\mathcal F_q, \mathcal E_q\right].
\end{align*}
For every $i\in S'^{(q)}$, the event
\[
\left\{t_q\geq\underline{\tau_q},\ 
\widehat\mu_i^q\geq\widehat\mu_1^q\right\}
\]
is contained in the event that the empirical mean of arm $i$ exceeds
that of arm $1$ at some common sample size $t_q\geq\underline{\tau_q}$.  Hence
Lemma~\ref{lem:adaptiveHoeffding} gives
\[
\mathbb P\left(
 t_q\geq\underline{\tau_q},\ 
 \widehat\mu_i^q\geq\widehat\mu_1^q
 \mid\mathcal F_q, \mathcal{E}_q
\right)
\leq \exp(-\underline{\tau_q}\Delta_i^2).
\]
It follows that
\[
\mathbb E\left[N_q\mathbf 1\{t_q\geq\underline{\tau_q}\}
\,\middle|\,\mathcal F_q, \mathcal{E}_q \right]
\leq |S'^{(q)}|\exp(-\underline{\tau_q}\Delta_{i_q}^2),
\]
and therefore
\begin{equation}\label{eq:phase-elimination-stochastic}
\mathbb P\left(\overline{\mathcal{E}}_{q+1}
\,\middle|\,\mathcal F_q, \mathcal{E}_q\right)
\leq
L\exp\left(-\frac{\underline{\tau_q}}{2}\right)
+3\exp(-\underline{\tau_q}\Delta_{i_q}^2).
\end{equation}

It remains to lower-bound the two exponents.  Since $f$ is increasing,
$f(d_{\ell,(1)}),\ldots,f(d_{\ell,(K)})$ is nonincreasing.  Thus
\[
\sum_{i=1}^{|S^{(q)}|}f(d_{\ell,(i)})
\leq 4\sum_{i=1}^{i_q}f(d_{\ell,(i)}).
\]
Consequently,
\begin{align*}
\underline{\tau_q}\Delta_{i_q}^2
&=\min_{\ell\in[L]}
\frac{C_\ell\Delta_{i_q}^2}{\log_2(K) \sum_{i=1}^{|S^{(q)}|}f(d_{\ell,(i)})}\\
&\geq \frac{1}{4\log_2(K)}
\min_{\ell\in[L]}
\frac{C_\ell}{\frac{1}{\Delta_{i_q}^2}\sum_{i=1}^{i_q}f(d_{\ell,(i)})}\\
&\geq \frac{\gamma^{\mathrm{sto}}(Q)}{4\log_2(K)}.
\end{align*}
Because the rewards lie in $[0,1]$, $\Delta_{i_q}\leq 1$, so also
\[
\underline{\tau_q}\geq \underline{\tau_q} \Delta_{i_q}^2
\geq \frac{\gamma^{\mathrm{sto}}(Q)}{4\log_2(K)}.
\]
Substituting these inequalities into \eqref{eq:phase-elimination-stochastic} yields
\begin{align*}
\mathbb P\left(\overline{\mathcal{E}}_{q+1}
\,\middle|\,\mathcal F_r\right)
&\leq
L\exp\left(-\frac{\gamma^{\mathrm{sto}}(Q)}{8\log_2(K)}\right)
+3\exp\left(-\frac{\gamma^{\mathrm{sto}}(Q)}{4\log_2(K)}\right)\\
&\leq
(L+3)\exp\left(-\frac{\gamma^{\mathrm{sto}}(Q)}{8\log_2(K)}\right).
\end{align*}

When $n_q=2$, there is only one competing arm.  Apply the adaptive
Hoeffding lemma directly, which also yields:
\begin{align*}
\mathbb P\left(\text{arm $1$ is eliminated in phase $\log_2(K)$}
\,\middle|\,\mathcal F_r\right)\leq
(L+3)\exp\left(-\frac{\gamma^{\mathrm{sto}}(Q)}{8\log_2(K)}\right).
\end{align*}

Summing over all phases yields the final result.
\end{proof}

\begin{proof}[Proof of Lemma \ref{lem:LBnbsamples}]

We have:
\begin{align}
\mathbb{P}\left(t_q\leq \underline{\tau_q}|\mathcal{F}_q \right)\leq& \mathbb{P}\left(\exists \ell \text{ s.t. the type $\ell$ ressource is exhausted after } \underline{\tau_t} \text{ samples}|\mathcal{F}_q \right).\notag\\
\leq& L\mathbb{P}\left(\text{type $\ell$ ressource is exhausted after } \underline{\tau_t} \text{ samples}|\mathcal{F}_q \right).\notag
\end{align}
Denote $d_\ell^{i,t}$ the ressource consumption of ressource type $\ell$ by arm $i$ on iteration $t$. We have:
\begin{align}
  \mathbb{P}\left(\text{ type $\ell$  exhausted at } \underline{\tau_q} \text{ samples}|\mathcal{F}_q \right) &=    \mathbb{P}\left(\sum_{t=\tau_{t-1}}^{\underline{\tau_q}+\tau_{q-1}}\sum_{i \in S^{(q)}}d_\ell^{i,t} \geq \frac{C_\ell}{\log_2(K)}|\mathcal{F}_q \right)  \notag\\
  &\hspace{-5cm}=  \mathbb{P}\left(\sum_{t=\tau_{q-1}}^{\underline{\tau}_q+\tau_{q-1}}\sum_{i \in S^{(q)}}d_\ell^{i,t} \geq \sum_{\tau=1}^{\underline{\tau}_q}\frac{C_\ell}{\log_2(K)\min_\ell \frac{C_\ell}{\log_2(K)\sum_{j=1}^{|S^{(q)}|}f\left(d_{\ell,(j)}\right)}}|\mathcal{F}_q \right)\notag\\
   &\hspace{-5cm}\leq   \mathbb{P}\left(\sum_{t=\tau_{r-1}}^{\underline{\tau_q}+\tau_{q-1}}\sum_{i \in S^{(q)}}d_\ell^{i,t} \geq \underline{\tau_q}\sum_{j=1}^{|S^{(q)}|}f\left(d_{\ell,(j)}\right)|\mathcal{F}_q \right)
\end{align}
By Lemma \ref{lem:lowerboundnumsamples1},
\[
  \mathbb{P}\left(\text{ type $\ell$  exhausted at } \underline{\tau_q} \text{ samples}|\mathcal{F}_q \right)\leq \exp(-  \underline{\tau_q}/2).
\]
A union bound over the $L$ possible resources gives the lemma.
\end{proof}

\begin{proof}[Proof of Lemma \ref{lem:adaptiveHoeffding}]
    By Hoeffding's lemma, for every $\lambda \in \mathbb{R}$,
    \[
        \mathbb{E}\left[
            \exp\left(\lambda(Y_i-\mathbb{E}[Y_i])\right)
        \right]
        \leq \exp\left(\frac{\lambda^2}{8}\right)
    \]
    and
    \[
        \mathbb{E}\left[
            \exp\left(-\lambda(X_i-\mathbb{E}[X_i])\right)
        \right]
        \leq \exp\left(\frac{\lambda^2}{8}\right).
    \]
    Since the two sequences are mutually independent, it follows that
    \[
    \begin{aligned}
        &\mathbb{E}\left[
            \exp\left(
                \lambda\bigl(
                    (Y_i-\mathbb{E}[Y_i])
                    -(X_i-\mathbb{E}[X_i])
                \bigr)
            \right)
        \right] \\
        &\qquad =
        \mathbb{E}\left[
            \exp\left(\lambda(Y_i-\mathbb{E}[Y_i])\right)
        \right]
        \mathbb{E}\left[
            \exp\left(-\lambda(X_i-\mathbb{E}[X_i])\right)
        \right]
        \leq \exp\left(\frac{\lambda^2}{4}\right).
    \end{aligned}
    \]
    Therefore, with respect to the natural filtration,
    \[
        M_t
        =
        \exp\left(
            \lambda
            \sum_{i=1}^t
            \bigl(
                (Y_i-\mathbb{E}[Y_i])
                -(X_i-\mathbb{E}[X_i])
            \bigr)
            -\frac{t\lambda^2}{4}
        \right),
        \qquad M_0=1,
    \]
    is a nonnegative supermartingale.

    If $\overline{X}_t\leq\overline{Y}_t$, then
    \[
    \begin{aligned}
        \sum_{i=1}^t
        \bigl(
            (Y_i-\mathbb{E}[Y_i])
            -(X_i-\mathbb{E}[X_i])
        \bigr)
        &=
        \sum_{i=1}^t(Y_i-X_i)
        +t\bigl(\mathbb{E}[X_1]-\mathbb{E}[Y_1]\bigr) \\
        &\geq t\Delta.
    \end{aligned}
    \]
    Choosing $\lambda=2\Delta$, we obtain, on this event,
    \[
        M_t
        \geq
        \exp\left(
            2\Delta\cdot t\Delta
            -\frac{t(2\Delta)^2}{4}
        \right)
        =
        \exp\left(t\Delta^2\right).
    \]
    Hence, for every $t\geq\tau$,
    \[
        \left\{\overline{X}_t\leq\overline{Y}_t\right\}
        \subseteq
        \left\{M_t\geq\exp(\tau\Delta^2)\right\}.
    \]
    Consequently,
    \[
    \begin{aligned}
        \mathbb{P}\left(
            \exists t\in\{\tau,\ldots,T\}:
            \overline{X}_t\leq\overline{Y}_t
        \right)
        &\leq
        \mathbb{P}\left(
            \max_{0\leq t\leq T}M_t
            \geq\exp(\tau\Delta^2)
        \right) \\
        &\leq \exp(-\tau\Delta^2),
    \end{aligned}
    \]
    where the last inequality follows from Ville's inequality.
\end{proof}
\begin{proof}[Proof of Lemma \ref{lem:lowerboundnumsamples1}] The case $\sum_{i=1}^m f(d_i)>m$ is trivial as the values lie in $[0;1]$. Let us consider the case $\sum_{i=1}^m f(d_i)\leq m$
    By Chernoff's inequality, 
    \begin{align}\label{eq:chernoff1}
        \mathbb{P}\left(
    \sum_{s=1}^{N}\sum_{i=1}^{m}X_{i,s}
    \geq N\sum_{i=1}^{m}f(d_i)
\right)\leq \inf_{\lambda\geq 0}\mathbb E \left[\exp\left(\lambda\sum_{s=1}^{N}\sum_{i=1}^{m}X_{i,s}\right)\right]\exp\left(-\lambda N\sum_{i=1}^{m}f(d_i)\right)
    \end{align}

    For each $X_{i,s}$, by convexity,
    \[
    \mathbb{E}\left[\exp\left(\lambda X_{i,s}\right)\right]\leq 1-d_i+ d_i e^\lambda= \exp\left(\log(1-d_i+ d_i e^\lambda)\right).
    \]
    Combining this with the independence of the $X_{i,s}$ yields:
    \[
    E \left[\exp\left(\lambda\sum_{s=1}^{N}\sum_{i=1}^{m}X_{i,s}\right)\right] \leq \exp \left(N \sum_{i=1}^m \log(1-d_i+ d_i e^\lambda))\right).
    \]
    Injecting in Equation \ref{eq:chernoff1}:
        \begin{align}\label{eq:chernoff1}
        \mathbb{P}\left(
    \sum_{s=1}^{N}\sum_{i=1}^{m}X_{i,s}
    \geq N\sum_{i=1}^{m}f(d_i)
\right)\leq \exp\left(-N\sup_{\lambda \geq 0}\sum_{i=1}^m \underbrace{\lambda f(d_i)-\log(1-d_i+ d_i e^\lambda)}_{g_i(\lambda)}\right)
    \end{align}
It remains to show  $\sup_{\lambda \geq 0}\sum_{i=1}^m g_i(\lambda)\geq 1/2$.

Consider first an index $i$ such that $f(d_i)<1$. Since $f(d_i)>d_i$, the function $g_i$ is maximized at
\[
\lambda_i^\star
:=
\log\left(
    \frac{f(d_i)(1-d_i)}
         {d_i(1-f(d_i))}
\right)>0,
\]
and the value of the maximum is
\[
g_i(\lambda_i^\star)
=
\kl(f(d_i),d_i),
\]
where $\kl$ denotes the Bernoulli relative entropy. Moreover,
$g_i$ is nondecreasing on $[0,\lambda_i^\star]$.

Suppose that there is at least one index such that $f(d_i)<1$, and choose
\[
j\in\argmin_{\{i:f(d_i)<1\}}\lambda_i^\star.
\]
For every $i$ such that $f(d_i)<1$, we have
$\lambda_j^\star\leq\lambda_i^\star$, and therefore
\[
g_i(\lambda_j^\star)\geq g_i(0)=0.
\]
For every $i$ such that $f(d_i)\geq1$, we have
\[
g_i'(\lambda)
=
f(d_i)-
\frac{d_i e^\lambda}
     {1-d_i+d_i e^\lambda}
\geq0,
\]
and hence again
\[
g_i(\lambda_j^\star)\geq g_i(0)=0.
\]
It follows that
\[
\sup_{\lambda \geq 0}\sum_{i=1}^m g_i(\lambda)\geq \sum_{i=1}^m g_i(\lambda_j^\star)
\geq
g_j(\lambda_j^\star)
=
\kl(f(d_j),d_j).
\]

Since $f(d_j)<1$, necessarily $d_j<e^{-2}$. As shown in the claim,
\[
\kl(f(d_j),d_j)\geq1/2.
\]
Consequently,
\[
\sup_{\lambda \geq 0}\sum_{i=1}^m g_i(\lambda)\geq 1/2.
\]

It remains to consider the case in which $f(d_i)\geq1$ for every
$i$. Since $\sum_{i=1}^mf(d_i)\leq m$, this forces $f(d_i)=1$ for every $i$, and hence
$d_i=e^{-2}$ for every $i$. In this case,
\[
\begin{aligned}
\sup_{\lambda \geq 0}\sum_{i=1}^m g_i(\lambda)
&\geq
\lim_{\lambda\to\infty}
\sum_{i=1}^{m}
\left[
\lambda-\log(1-d_i+d_i e^\lambda)
\right]\\
&=
\sum_{i=1}^{m}\log\frac1{d_i}
=
2m
\geq1.
\end{aligned}
\]

Thus, in all cases,
\[
\sup_{\lambda \geq 0}\sum_{i=1}^m g_i(\lambda)\geq 1/2.
\]
Substituting this into the Chernoff bound gives
\[
\mathbb{P}\left(
    \sum_{s=1}^{N}\sum_{i=1}^{m}X_{i,s}
    \geq
    N\sum_{i=1}^{m}f(d_i)
\right)
\leq e^{-N/2}.
\]
\end{proof}

\section{Proof of the main Theorem}

 Here, we derive Theorem 4.1 from the main structural proposition.

For each phase $q$, denote $\mathcal{A}_q$ the event that the surviving arms are collectively well-concentrated around the top of the reward distribution:
\[
\mathcal{A}_q:= \left\{\frac{1}{n_{q}}\sum_{i \in S_{q}} \Delta_i \leq 3\Delta_{2n_{q}}.\right\}
\]
Informally, $\mathcal{A}_q$ asserts that the average suboptimality of the surviving set at phase $q$ is within a constant factor of $\Delta_{2n_q}$,
the gap of the arm at rank $2n_q$ in the overall reward ordering.

Under $\mathcal{A}_q$, each round-robin cycle consumes at most
$3n_q\Delta_{2n_q}$ expected regret. Recall the definition of  $\underline{\tau_q}$, which we will show to be a conservative lower
bound on the number of cycles completed in phase $q$:
\[
\underline{\tau_q}=\min\left(\frac{T}{n_q\log_2 K},\ \frac{B^{(q)}}{6n_q\Delta_{2n_q}}\right).
\]

Let us reformulate the main structural result (Proposition 4.1) with these notations.
\begin{proposition}\label{prop:lowerboundAq}
    If for all $q'$, $\log 4e\leq\frac{1}{8}
   \underline \tau_{q'}\left(\frac{\Delta_{\frac{n_{q'}}{4}+1}}{\log_2(K)}\right)^2$, then
    \[
    \mathbb{P}\left(\overline{\bigcap_q \mathcal{A}_q}\right)\leq 3\log_2(K)^2\exp\left(-\frac{1}{8\log_2(K)^2}\min_q\underline{\tau}_{q}\Delta_{\frac{n_{q}}{4}+1}^2\right).
    \]
\end{proposition}
 Note that this is the same proposition as in the main text, since $
\Gamma_{2,R}=\min_{q\in [\log_2(K)]}\underline{\tau}_q\Delta_{\frac{n_q}{4}+1}^2.
$ The above proposition is proved in Section \ref{sec:prooflowerboundAq}. The main consequence of event $\mathcal{A}_q$, is that, provided it holds, then the number of pulls per arm in the phase is lower bounded.
\begin{proposition}\label{lem:LBtauq}
    For each phase $q$,
    \[
    \mathbb{P}\left(t_q<\underline{\tau_q}|\mathcal{A}_q, \mathcal{F}_q\right)\leq \exp\left(-18\underline{\tau_q}\Delta_{2n_q}^2\right)
    \]
    
\end{proposition}

\noindent Proposition \ref{lem:LBtauq} states that, conditional on arm quality event $\mathcal{A}_q$, the phase completes at least $\underline{\tau_q}$ cycles with high probability. The only way fewer cycles are completed is if the regret budget $B^{(q)}$ is exhausted early due to upward deviations of the realized consumption above its expectation, an event whose probability decays exponentially in $\underline{\tau_q}\Delta_{2n_q}^2$.

Once this is established, standard SH analysis argument yield the following proposition. Define event:
\[
\mathcal{E}_{q}:= \left\{1 \in S^{(q)}\right\}.
\]
\begin{proposition}\label{prop:bestarmeliminated}
    For each phase $q$, 
    \[
    \mathbb{P}\left(\overline{\mathcal{E}}_{q+1}|\mathcal{E}_{q}, \mathcal{A}_q, \mathcal{F}_q\right)\leq 4\exp\left(-
\underline{\tau_q}\Delta_{\frac{n_q}{4}+1}^2\right).
    \]
\end{proposition}
Suppose that for all $q'$, $\log 4e\leq\frac{1}{8}
   \underline \tau_{q'}\left(\frac{\Delta_{\frac{n_{q'}}{4}+1}}{\log_2(K)}\right)^2$. To prove the main theorem, notice first that $\mathcal{E}_{\log_2(K)+1}= \{\phi=1\}$. Then:
\begin{align*}
    \mathbb{P}\left(\phi \neq 1\right)=& \mathbb{P}\left(\phi \neq 1, \bigcap_q \mathcal{A}_q\right)+\mathbb{P}\left(\phi \neq 1, \overline{\bigcap_q \mathcal{A}_q}\right)\\
    \leq& \mathbb{P}\left(\phi \neq 1, \bigcap_q \mathcal{A}_q\right)+\mathbb{P}\left(\overline{\bigcap_q \mathcal{A}_q}\right)
\end{align*}
We can upper bound the term on the right via Proposition \ref{prop:lowerboundAq}. For the term on the left:
\begin{align*}
    \mathbb{P}\left(\phi \neq 1, \bigcap_q \mathcal{A}_q\right)=&\mathbb{P}\left(\overline{\mathcal{E}_{\log_2(K)+1}}, \bigcap_q \mathcal{A}_q\right)\\
    \leq&\sum_{q=1}^{\log_2(K)} \mathbb{P}\left(\overline{\mathcal{E}_{q+1}},\mathcal{E}_{q},  \mathcal{A}_q\right)\\
    \leq &\sum_{q=1}^{\log_2(K)} \mathbb{P}\left(\overline{\mathcal{E}_{q+1}}|\mathcal{E}_{q},  \mathcal{A}_q\right)
\end{align*}
Then, Proposition \ref{prop:bestarmeliminated} yields:
\begin{align*}
    \mathbb{P}\left(\phi \neq 1, \bigcap_q \mathcal{A}_q\right)\leq&\sum_{q=1}^{\log_2(K)} 4\exp\left(-
\underline{\tau_q}\Delta_{\frac{n_q}{4}+1}^2\right)\\
\leq &4\log_2(K)\exp\left(-\min_q
\underline{\tau_q}\Delta_{\frac{n_q}{4}+1}^2\right)\\
\leq &4\log_2(K)^2\exp\left(-\frac{1}{8\log_2(K)^2}\min_q
\underline{\tau_q}\Delta_{\frac{n_q}{4}+1}^2\right)
\end{align*}
Combining the two bounds:
\[
\mathbb{P}\left(\phi \neq 1\right)\leq 7\log_2(K)^2\exp\left(-\frac{1}{8\log_2(K)^2}\min_q
\underline{\tau_q}\Delta_{\frac{n_q}{4}+1}^2\right).
\]

\subsection{Proof of Proposition \ref{prop:lowerboundAq}}\label{sec:prooflowerboundAq}

For each phase $q$, we will see that event $A_{q'}$ holds due to a two stage chain of event. First, few arms in the surviving set have gap small than $\Delta_{2n_{q'}}$. In this case, most of them survive. Then, once the average of the gap a significant portion of the arms is order dominated by $\Delta_{2n_{q'}}$, it remains true for the subsequent phases, and that until phase $q'$. Then, when everything goes well in the chain of event, after phase $q'-2$, the event that the average gap of a lot of arms is order dominated by $\Delta_{2n_{q'}}$ happens deterministically.

Before giving the full proof, we introduce each subevent in that chain and study how they imply each other.\\

For each phase pair $q' > q + 1$, define the good-arm conservation event:
\[
\mathcal{C}_{q',q}:= \left\{C_{q}(\Delta_{2n_{q'}})\geq 
C_1(\Delta_{2n_{q'}})\left(1-\frac{q}{2\log_2(K)}\right) \right\},
\]
where $C_q(\Delta_{2n_{q'}})$ denotes the number of arms in $S^{(q)}$ with
suboptimality gap at most $\Delta_{2n_{q'}}$. $\mathcal{C}_{q',q}$
ensures that a controlled fraction of the near-optimal arms present at
phase 1 remain in the surviving set at phase $q$ and, essentially,
that elimination has not discarded too many good arms. The event:
\[
\mathcal{T}_{q',q}:=\left\{\Delta_{\frac{n_{q}}{4}+1}(q)\leq 
2\Delta_{2n_{q'}}\right\},
\]
where $\Delta_{\frac{n_q}{4}+1}(q)$ denotes the suboptimality gap of
the $\left(\frac{n_q}{4}+1\right)$-th best arm within the surviving
set $S^{(q)}$, ensures the surviving set has concentrated sufficiently
around the top arms relative to the reference gap $\Delta_{2n_{q'}}$.
Together, $\mathcal{C}_{q',q}$ counts whether enough good arms remain,
while $\mathcal{T}_{q',q}$ measures whether the survivors are
collectively good enough.

\begin{proposition}\label{lem:verygoodarmconservation}
    For each phase pair $q'>q+1$ with $n_q \geq 8$, if 
    $\log(2e\log K)\leq \frac{\Delta_{\frac{n_q}{4}+1}^2\underline{\tau_q}}{16}$,
    \[
\mathbb{P}\left(\overline{\mathcal{C}}_{q',q+1}\mid\mathcal{C}_{q',q}, 
\mathcal{A}_q, \mathcal{F}_q, \overline{\mathcal{T}}_{q',q}\right)\leq 
3\exp\left(-\frac{1}{16}\underline{\tau_q}\Delta_{\frac{n_q}{4}+1}^2\right).
    \]
\end{proposition}
Proposition~\ref{lem:verygoodarmconservation} establishes that good arm
conservation propagates from phase $q$ to $q+1$ in the regime where
$\mathcal{C}_{q',q}$ holds but $\mathcal{T}_{q',q}$ does not, provided
the phase length $\underline{\tau_q}$ is sufficient relative to
$\Delta_{\frac{n_q}{4}+1}$. That is, when the surviving set has
not yet concentrated around the top arms, elimination does not discard
good arms with high probability.

For each phase pair $q'>q$, define the event:
\[
\mathcal{M}_{q',q}:=\left\{\frac{2}{n_{q}}\sum_{i=1}^{n_{q}/2}
\Delta_i(q)\leq \left(2+\frac{q}{\log_2(K)}\right)\Delta_{2n_{q'}},\ 
\Delta_{\frac{n_q}{4}+1}\leq 2\Delta_{2n_{q'}}\right\}.
\]
The first half of $\mathcal{M}_{q',q}$ bounds how good the top half of survivors are on average relative to the reference gap $\Delta_{2n_{q'}}$. Notably, since each elimination retains the top half of $S^{(q)}$ to form $S^{(q+1)}$, $\mathcal{M}_{q',q}$ directly characterizes the arm quality of the subsequent phase. For the second half of $\mathcal{M}_{q',q}$, note that it is implied by $\mathcal{T}_{q',q}$, and then remains true once it holds for a phase, as $n_q$ is decreasing.

\begin{proposition}\label{prop:LBtoparmdecay}
    For each phase $q'>q$ with $n_q \geq 8$, if $\log 4e\leq\frac{1}{8}
   \underline \tau_q\left(\frac{\Delta_{\frac{n_q}{4}+1}}{\log_2(K)}\right)^2$,
    \[
    \mathbb{P}\left(\overline{\mathcal{M}}_{q',q+1}|\mathcal{M}_{q',q}, \mathcal{A}_q, \mathcal{F}_q\right)\leq 2\exp\left(-\frac{1}{8\log_2(K)^2}\underline{\tau_q}\Delta_{\frac{n_q}{4}+1}^2\right)
    \]
    and for each $q$, $n_q \geq 4$, if $\log 4e\leq\frac{1}{8}
   \underline \tau_q\left(\frac{\Delta_{\frac{n_q}{4}+1}}{\log_2(K)}\right)^2$,
      \[
    \mathbb{P}\left(\overline{\mathcal{A}}_{q+1}|\mathcal{M}_{q+1,q}, \mathcal{A}_q, \mathcal{F}_q\right)\leq 2\exp\left(-\frac{1}{8\log_2(K)^2}\underline{\tau_q}\Delta_{\frac{n_q}{4}+1}^2\right).
    \]
\end{proposition}
Proposition~\ref{prop:LBtoparmdecay} handles the post-transition
regime. It establishes two results: first, that $\mathcal{M}_{q',q}$
propagates to $\mathcal{M}_{q',q+1}$ with high probability; second,
that given $\mathcal{M}_{q+1,q}$, the arm quality event
$\mathcal{A}_{q+1}$ holds with high probability.

\begin{proposition}\label{lem:transitionphase}
    For each phase $q'>q$ with $n_q \geq 8$, if $\log 4e\leq\frac{1}{8}
   \underline \tau_q\left(\frac{\Delta_{\frac{n_q}{4}+1}}{\log_2(K)}\right)^2$,
    \[
    \mathbb{P}\left(\overline{\mathcal{M}}_{q',q+1}|\mathcal{T}_{q',q}, \mathcal{A}_q, \mathcal{F}_q\right)\leq 2\exp\left(-\frac{1}{8\log_2(K)^2}\underline{\tau_q}\Delta_{\frac{n_q}{4}+1}^2\right).
    \]
\end{proposition}

Finally, we have the following deterministic claim.
\begin{claim}\label{claim:deterministicimplication}
    For any $q'\in [\log_2(K)]$, $\mathcal C_{q',q'-2}$ implies $\mathcal{T}_{q',q'-2}$.
\end{claim}
\begin{proof}[Proof of claim \ref{claim:deterministicimplication}]
    
There are at least \(2n_{q'}\) arms whose gaps are no larger than
\(\Delta_{2n_{q'}}\). Hence
\[
C_1(\Delta_{2n_{q'}})\geq 2n_{q'}.
\]
On \(\mathcal C_{q',q}\), it follows that
\[
C_q(\Delta_{2n_{q'}})
\geq
C_1(\Delta_{2n_{q'}})
\left(1-\frac{q}{2\log_2 K}\right)
\geq
2n_{q'}
\left(1-\frac{q}{2\log_2 K}\right).
\]
By integrality, this implies $C_q(\Delta_{2n_{q'}})\geq n_{q'}+1$.
On the other hand, on \(\overline{\mathcal{T}}_{q',q}\),
\[
\Delta_{\frac{n_q}{4}+1}(q)
>
2\Delta_{2n_{q'}}.
\]
Because the arms in \(S^{(q)}\) are also ordered increasingly according
to their gaps, this means that at most \(n_q/4\) arms in \(S^{(q)}\)
have gap at most \(2\Delta_{2n_{q'}}\). In particular,
\[
C_q(\Delta_{2n_{q'}})
\leq
C_q(2\Delta_{2n_{q'}})
\leq
\frac{n_q}{4}.
\]
Combining (1) and (2), we obtain the deterministic inclusion
\[
\mathcal C_{q',q}\cap\overline{\mathcal{T}}_{q',q}
\subseteq
\left\{
n_{q'}+1
\leq
\frac{n_q}{4}
\right\}.
\]

This implies that if $\mathcal C_{q',q'-2}$ holds, then so does $\mathcal{T}_{q',q'-2}$.  
\end{proof}

\begin{proof}[Proof of Proposition \ref{prop:lowerboundAq}]

First, for every \(q'\), the event \(\mathcal C_{q',1}\) holds
by definition. Since \(S^{(1)}\) contains all the arms,  $\mathcal{A}_1$ and $\mathcal{A}_2$ as well as $\mathcal{M}_{2,1}$ hold deterministically.

Fix now \(q'\geq 2\), and suppose that
\[
\bigcap_{q<q'}\mathcal A_q
\]
holds. We follow the evolution of the arms whose gaps are at most
\(\Delta_{2n_{q'}}\).

Starting from \(\mathcal C_{q',1}\), consider a phase \(q\leq q'-2\).
As long as \(\mathcal{T}_{q',q}\) does not hold, Proposition
\ref{lem:verygoodarmconservation} shows that
\(\mathcal C_{q',q}\) is propagated to \(\mathcal C_{q',q+1}\), except
with probability at most
\[
3\exp\left(
-\frac{1}{16}\underline{\tau_q}
\Delta_{\frac{n_q}{4}+1}^{2}
\right)\leq 3\exp\left(
-\frac{1}{8\log_2(K)^2}\underline{\tau_q}
\Delta_{\frac{n_q}{4}+1}^{2}
\right).
\]
By Claim~\ref{claim:deterministicimplication}, it is impossible to
reach phase \(q'-2\) while both
\(\mathcal C_{q',q'-2}\) and \(\overline{\mathcal{T}}_{q',q'-2}\) hold.
Consequently, provided none of the preceding conservation steps
fails, \(\mathcal{T}_{q',q}\) must hold for some \(q\leq q'-2\).

At the first such phase, Proposition~\ref{lem:transitionphase} gives
\(\mathcal M_{q',q+1}\), except with probability at most
\[
2\exp\left(
-\frac{1}{8\log_2(K)^2}\underline{\tau_q}
\Delta_{\frac{n_q}{4}+1}^{2}
\right)\leq 3\exp\left(
-\frac{1}{8\log_2(K)^2}\underline{\tau_q}
\Delta_{\frac{n_q}{4}+1}^{2}
\right).
\]
Once \(\mathcal M_{q',q+1}\) holds, the first part of Proposition
\ref{prop:LBtoparmdecay} propagates it through the subsequent phases:
\[
\mathcal M_{q',q+1}
\Longrightarrow
\mathcal M_{q',q+2}
\Longrightarrow\cdots\Longrightarrow
\mathcal M_{q',q'-1},
\]
with the failure probability at each phase bounded by
\[
2\exp\left(
-\frac{1}{8\log_2(K)^2}\underline{\tau_q}
\Delta_{\frac{n_q}{4}+1}^{2}
\right)\leq 3\exp\left(
-\frac{1}{8\log_2(K)^2}\underline{\tau_q}
\Delta_{\frac{n_q}{4}+1}^{2}
\right).
\]
Finally, applying the second part of Proposition
\ref{prop:LBtoparmdecay} at phase \(q'-1\) yields
\(\mathcal A_{q'}\) from \(\mathcal M_{q',q'-1}\), with the same
probability bound.

At every phase, conditional on the past, exactly one of the preceding three steps is relevant: conservation of \(\mathcal C_{q',q}\),
transition from \(\mathcal{T}_{q',q}\) to \(\mathcal M_{q',q+1}\), or propagation
of \(\mathcal M_{q',q}\). Therefore, after conditioning on
\(\mathcal F_q\) and using the tower property, the probability that
the construction leading to \(\mathcal A_{q'}\) first fails at phase
\(q\) is at most
\[
3\exp\left(
-\frac{1}{8\log_2(K)^2}\underline{\tau_q}
\Delta_{\frac{n_q}{4}+1}^{2}
\right).
\]
A union bound over the phases preceding \(q'\) consequently gives
\[
\mathbb P\left(
\overline{\mathcal A}_{q'}
\cap\bigcap_{q<q'}\mathcal A_q
\right)
\leq
3\sum_{q<q'}
\exp\left(
-\frac{1}{8\log_2(K)^2}\underline{\tau_q}
\Delta_{\frac{n_q}{4}+1}^{2}
\right).
\]

Applying the preceding estimate and a union bound over \(q'\), we
obtain
\[
\begin{aligned}
\mathbb P\left(
\overline{\bigcap_{q'}\mathcal A_{q'}}
\right)
&\leq
3\sum_{q'}\sum_{q<q'}
\exp\left(
-\frac{1}{8\log_2(K)^2}\underline{\tau_q}
\Delta_{\frac{n_q}{4}+1}^{2}
\right)                                                     \\
&\leq
3\log_2(K)^2
\exp\left(
-\frac{1}{8\log_2(K)^2}
\min_q
\left\{
\underline{\tau_q}
\Delta_{\frac{n_q}{4}+1}^{2}
\right\}
\right).
\end{aligned}
\]
Taking \(q\) in the statement to be a phase at which
\(\underline{\tau_q}\Delta_{\frac{n_q}{4}+1}^{2}\) is smallest gives
the claimed bound.

\end{proof}
\subsection{Proof of the propositions}

We start with the following useful claim, which can be derived from Ville's maximal inequality, and a direct consequence of Corollay 3 in \cite{Howard_2020}.
\begin{claim}\label{lem:anytimeHoeff}
    Let $(X_t)_{t \in \mathbb{N}}$ be a sequence of independent random variables in $[0,1]$. For any fixed threshold $a\geq 0$ and any $m$,
    \[
    \mathbb{P}\left(\exists t \geq m : \sum_{i\leq t}X_i\geq \sum_{i\leq t}\mathbb{E}[X_i]+ ta\right)\leq \exp(-2m a^2).
    \]
\end{claim}
Note that a lower tail version of that claim can be obtained by applying the result to $1-X_i$.
\begin{proof}[Proof of Proposition \ref{lem:LBtauq}]
Let $I_s\in S^{(q)}$ be the arm pulled at the $s$-th pull of the phase, and let $X_s\in[0,1]$ be the corresponding reward. Define $M_s$ as the deviation from expectation on arm pull $s$, i.e:
\[
M_s:= \sum_{u=1}^s\mu_{I_u}-X_u.
\]

Since the algorithm keeps a margin to complete round robin cycles, if $t_q< \underline{\tau_q}$, this implies that the regret budget (minus the margin) was exhausted at the end of some round-robin cycle. At the end of a complete round-robin cycle, every arm in $S^{(q)}$ has been pulled equally often, so, under event $\mathcal{A}_q$, the total expected regret consumption at a pull $s$ which is the end of a round robin cycle is at most $s(3\Delta_{2n_q}-\Delta_0)\leq  3s\Delta_{2n_q}$. Therefore, on the event $\mathcal{A}_q$, if $t_q\leq \underline{\tau_q}$, then for some $s \leq n_q \underline{\tau_q}$:
\[
M_s \geq 3 \underline{\tau_q} n_q \Delta_{2n_q}.
\]
$M_s$ is a martingale with respect to the filtration generated by $\mathcal{F}_q$ and the rewards observed during phase $q$, with bounded increments. By maximal Azuma-Hoeffding inequality:
\begin{align*}
    \mathbb{P}\left(\max_{1\leq s\leq n_q\underline{\tau_q}}M_s \geq 3 \underline{\tau_q} n_q\Delta_{2n_q}\right) \leq \exp\left(-18 \underline{\tau_q} n_q \Delta_{2n_q}^2\right).
\end{align*}

\end{proof}
\begin{proof}[Proof of Proposition \ref{lem:verygoodarmconservation}]
    The proof of this proposition relies on the following lemma, whose proof we delay:
    \begin{lemma}[Specialized anytime version of Lemma 9.1 from "Revisiting simple regret"]\label{lem:9.1revisited}
Suppose that, after $t$ pulls per arm, uniform sampling outputs the $K/2$ arms with the largest empirical means. Let $C_t(\epsilon)$ be the set of $\epsilon$-good
arms in the output, and let
\[
    g(\epsilon):=\left|\{i:\mu_i\ge \mu^\star-\epsilon\}\right|.
\]
Assume that $\frac{K}{2}>g(\epsilon)$. Let $\ell$ satisfy
\[
    g(\epsilon)\leq\ell<\frac{K}{2},
\]
and define
\[
    d_\ell:=\mu_{g(\epsilon)}-\mu_{\ell+1}.
\]
Then, for every integer $k\in\{1,\dots,g(\epsilon)\}$ and every
$\underline \tau \ge 1$,
\[
\mathbb{P}\left(
\exists t \geq \underline{\tau} \text{ s.t. }
\left|C_t( \epsilon)\right| \leq g(\epsilon)-k
\right)
\leq
\binom{g(\epsilon)}{k}
\exp\left(-k\frac{d_\ell^2\underline \tau}{2}\right)
+
\binom{K-\ell}{\frac{K}{2}-\ell+k}
\exp\left(-\left(\frac{K}{2}-\ell+k\right)\frac{d_\ell^2\underline \tau}{2}\right),
\]
with the convention that $\binom{a}{b}=0$ when $b>a$.
\end{lemma}
 We apply the above lemma with $\epsilon = \Delta_{2n_{q'}}$, $K=n_q $, $\ell=\frac{n_q}{4}$. Let us first derive the value of $d_\ell$ with that instantiation. By definition,
 \[
 d_\ell \geq \mu^*-\Delta_{2n_{q'}}-\mu_{\frac{n_q}{4}+1}(q)= \Delta_{\frac{n_q}{4}+1}(q)-\Delta_{2n_{q'}}.
 \]
 Event $\overline{\mathcal{T}}_{q',q}$ then implies:
  \[
 d_\ell \geq \frac{1}{2}\Delta_{\frac{n_q}{4}+1}(q) \geq \frac{1}{2}\Delta_{\frac{n_q}{4}+1}.
 \]
 
Then, for any $k \leq C_q(\Delta_{2n_{q'}})$,
    \begin{align*}
\mathbb{P}\left(
C_{q+1}(\Delta_{2n_{q'}})\leq C_{q}(\Delta_{2n_{q'}})-k
|\mathcal{C}_{q',q}, \mathcal{A}_q, \mathcal{F}_q, \overline{\mathcal{T}}_{q',q}\right)
\leq&
\binom{ C_{q}(\Delta_{2n_{q'}})}{k}
\exp\left(-k\frac{\Delta_{\frac{n_q}{4}+1}^2\underline \tau_q}{8}\right)\\
&+
\binom{\frac{3}{4}n_q}{\frac{n_q}{4}+k}
\exp\left(-(\frac{n_q}{4}+k)\frac{\Delta_{\frac{n_q}{4}+1}^2\underline \tau_q}{8}\right),\\
&+\mathbb{P}\left(t_q\leq \underline{\tau_q}|\mathcal{C}_{q',q}, \mathcal{A}_q, \mathcal{F}_q, \overline{\mathcal{T}}_{q',q}\right)
    \end{align*}
Instantiating $k= \lceil \frac{1}{2\log_2(K)}C_q(\Delta_{2n_{q'}})\rceil$ and the inequality $\binom{ x}{y} \leq \left(\frac{ex}{y}\right)^y$, this yields:
\[
\begin{aligned}
&\mathbb{P}\left(
C_{q+1}(\Delta_{2n_{q'}})\leq C_{q}(\Delta_{2n_{q'}})-k
|\mathcal{C}_{q',q}, \mathcal{A}_q, \mathcal{F}_q, \overline{\mathcal{T}}_{q',q}\right)
\\
&\leq
\exp\left[
\lceil\frac{C_q(\Delta_{2n_{q'}})}{2\log_2 K}\rceil
\left(
\log(2e\log K)
-
\frac{\Delta_{\frac{n_q}{4}+1}^2\underline \tau_q}{8}
\right)
\right]
\\
&\quad+
\exp\left[
\left(
\frac{n_q}{4}
+
\lceil\frac{C_q(\Delta_{2n_{q'}})}{2\log K}\rceil
\right)
\left(
\log(3e)
-
\frac{\Delta_{\frac{n_q}{4}+1}^2\underline \tau_q}{8}
\right)
\right]
+\mathbb{P}\left(t_q\leq \underline{\tau_q}|\mathcal{C}_{q',q}, \mathcal{A}_q, \mathcal{F}_q, \overline{\mathcal{T}}_{q',q}\right)
\end{aligned}
\]
Finally, the condition $\log(2e\log K)\leq \frac{\Delta_{\frac{n_q}{4}+1}^2\underline \tau_q}{16}$ together with Proposition \ref{lem:LBtauq} yields:
\[
\begin{aligned}
&\mathbb{P}\left(
C_{q+1}(\Delta_{2n_{q'}})\leq C_{q}(\Delta_{2n_{q'}})-k
|\mathcal{C}_{q',q}, \mathcal{A}_q, \mathcal{F}_q, \overline{\mathcal{T}}_{q',q}\right)\leq 3 \exp\left(
-\frac{\Delta_{\frac{n_q}{4}+1}^2\underline \tau_q}{16}
\right)
\end{aligned}
\]

\end{proof}

\begin{proof}[Proof of Proposition \ref{prop:LBtoparmdecay}]
    The proof relies on  a combination of Proposition \ref{lem:LBtauq} and  the following intermediate result whose proof is delayed:
    \begin{lemma}[Anytime bound on top-arms mean decay]\label{lem:meandecay} Consider a set of $K$ arm, sampled in a round robin fashion. At round $t$ (where each arm has been sampled $t$ times), denote $A(t)$ the set of $K/2$ arms with the largest empirical mean, ties broken at random.
For $1\leq s\leq K/2$, let
\[
    B_s:=\{1,\dots,s\}
\]
be the set of the true top $s$ arms, and let $B_s(t)$ be the set of the $s$ arms with largest true means inside $A(t)$, ties broken in favor of elements in $B_s$. Define
\[
    \mu_{B_s}:=\frac1s\sum_{i\in B_s}\mu_i,
    \qquad
    \mu_{B_s(t)}:=\frac1s\sum_{i\in B_s(t)}\mu_i .
\]
Then, for every $\epsilon>0$ and every $\underline \tau\geq 1$,
\[
\mathbb P\left(
    \exists t\geq \underline \tau:
    \mu_{B_s(t)}\leq \mu_{B_s}-\epsilon
\right)
\leq
\exp\left(
    s\log\frac{eK}{s}
    -
   \underline \tau\,s\epsilon^2
\right).
\]

\end{lemma}

To obtain the first bound, we instantiate the above lemma with $s=\frac{n_q}{4}$, $K=n_q$, $\epsilon= \frac{\Delta_{2n_{q'}}}{\log_2(K)}$ and $\underline{\tau}= \underline{\tau_q}$. We have:
\begin{align*}  \mathbb{P}\left(\overline{\mathcal{M}}_{q',q+1}|\mathcal{M}_{q',q}, \mathcal{A}_q,\mathcal{F}_q\right)\leq& \mathbb{P}\left(\exists t\geq \underline \tau_q:
    \mu_{B_{\frac{n_q}{4}}(t)}\leq \mu_{B_{\frac{n_q}{4}}}-\frac{\Delta_{2n_{q'}}}{\log_2(K)}|\mathcal{M}_{q',q}, \mathcal{A}_q,\mathcal{F}_q\right)\\
    &+\mathbb{P}\left(t_q\leq \tau_q|\mathcal{M}_{q',q}, \mathcal{A}_q,\mathcal{F}_q\right)
\end{align*}

We upper bound the term in the fist line via Lemma \ref{lem:meandecay} and the second term via Proposition \ref{lem:LBtauq}.
\begin{align*}  \mathbb{P}\left(\overline{\mathcal{M}}_{q',q+1}|\mathcal{M}_{q',q}, \mathcal{F}_q\right)\leq& \exp\left(  \frac{n_q}{4}\log 4e
    -
   \underline \tau_q\,\frac{n_q}{4}\left(\frac{\Delta_{2n_{q'}}}{\log_2(K)}\right)^2
\right)\\
    &+\exp\left(-18\underline{\tau_q}\Delta_{2n_q}^2\right)
\end{align*}
 Using the condition $\Delta_{\frac{n_q}{4}+1}\leq 2\Delta_{2n_{q'}}$ and the fact $\Delta_{\frac{n_q}{4}+1}\leq \Delta_{2n_{q}}$, we obtain
 \begin{align*}  \mathbb{P}\left(\overline{\mathcal{M}}_{q',q+1}|\mathcal{M}_{q',q}, \mathcal{F}_q\right)\leq& \exp\left(  \frac{n_q}{4}\log 4e
    -
   \underline \tau_q\,\frac{n_q}{16}\left(\frac{\Delta_{\frac{n_q}{4}+1}}{\log_2(K)}\right)^2
\right)\\
    &+\exp\left(-18\underline{\tau_q}\Delta_{\frac{n_q}{4}+1}^2\right)
\end{align*}
Finally, from the condition $\log 4e\leq\frac{1}{8}
   \underline \tau_q\left(\frac{\Delta_{\frac{n_q}{4}+1}}{\log_2(K)}\right)^2$, and $n_q\geq 8$ we obtain:
    \begin{align*}  \mathbb{P}\left(\overline{\mathcal{M}}_{q',q+1}|\mathcal{M}_{q',q}, \mathcal{F}_q\right)\leq& 
    2\exp\left(-\frac{1}{4\log_2(K)^2}\underline{\tau_q}\Delta_{\frac{n_q}{4}+1}^2\right)
\end{align*}
For the second bound, we repeat the proof with $s=\frac{n_q}{2}$ and use the fact that $\frac{q+1}{\log_2(K)}\leq 1$.
\end{proof}
\begin{proof}[Proof of Proposition \ref{lem:transitionphase}]
Note that event $\mathcal{T}_{q',q}$ automatically implies that
\[
\frac{2}{n_{q+1}}\sum_{i=1}^{n_{q+1}/2}\Delta_i(q)\leq 2\Delta_{2n_q'} \text{ and } \Delta_{\frac{n_{q}}{4}+1}\leq \Delta_{\frac{n_{q}}{4}+1}(q)\leq 2\Delta_{2n_{q'}}. 
\]
Therefore:

\begin{align*}  \mathbb{P}\left(\overline{\mathcal{M}}_{q',q+1}|\mathcal{T}_{q',q}, \mathcal{F}_q, \mathcal{A}_q\right)\leq& \mathbb{P}\left(\exists t\geq \underline \tau_q:
    \mu_{B_{\frac{n_q}{4}}(t)}\leq \mu_{B_{\frac{n_q}{4}}}-\frac{\Delta_{2n_{q'}}}{\log_2(K)}|\mathcal{T}_{q',q}, \mathcal{A}_q,\mathcal{F}_q\right)\\
    &+\mathbb{P}\left(t_q\leq \tau_q|\mathcal{T}_{q',q}, \mathcal{A}_q,\mathcal{F}_q\right)
\end{align*}
We can then proceed repeating the steps from the proof of \ref{prop:LBtoparmdecay}.
\end{proof}

\begin{proof}[Proof of Proposition \ref{prop:bestarmeliminated}]

First suppose that $n_q\geq 4$.  Put
\[
i_q:=\frac{n_q}{4}+1,
\]
and  define $S^{\prime(q)}$ as the set of arms in $S^{(q)}$, excluding the $\frac{1}{4}n_q$ arms with the largest mean.
Every arm $i\in S^{\prime(q)}$ has global reward rank at least
$i_q$, and hence
\[
\Delta_i\geq \Delta_{i_q}.
\]
Let
\[
N_q:=\sum_{i\in S'^{(q)}}
\mathbf 1\!\left\{\widehat\mu_i^q\geq\widehat\mu_1^q\right\}.
\]
 If the best arm is eliminated in phase $q$, it must be the case that at least half the arms of $S^{(q)}$ have their empirical average larger than its empirical average. In particular, this implies
\[
N_q\geq \frac{n_q}{4}=\frac{|S'^{(q)}|}{3}.
\]
Therefore,
\begin{align*}
\mathbb P\left(\overline{\mathcal{E}}_{q+1}
\,\middle|\,\mathcal F_q, \mathcal{E}_q,\mathcal{A}_q\right)\leq&
\mathbb P(t_q<\underline{\tau_q}\mid\mathcal \mathcal F_q, \mathcal{E}_q,\mathcal{A}_q)
+
\mathbb P\left(
N_q\geq \frac{|S'^{(q)}|}{3},\ t_q\geq\underline{\tau_q}
\,\middle|\,\mathcal F_q, \mathcal{E}_q,\mathcal{A}_q
\right)
\end{align*}
By Proposition \ref{lem:LBtauq},
    \[
    \mathbb{P}\left(t_q<\underline{\tau_q}|\mathcal{A}_q, \mathcal{F}_q\right)\leq \exp\left(-18\underline{\tau_q}\Delta_{2n_q}^2\right)
    \]
By Markov's inequality:

\begin{align*}
\mathbb P\left(
N_q\geq \frac{|S'^{(q)}|}{3},\ t_q\geq\underline{\tau_q}
\,\middle|\,\mathcal F_q, \mathcal{E}_q
\right)\leq&
\frac{3}{|S'^{(q)}|}
\mathbb E\left[N_q\mathbf 1\{t_q\geq\underline{\tau_q}\}
\,\middle|\,\mathcal F_q, \mathcal E_q\right].
\end{align*}
For every $i\in S'^{(q)}$, the event
\[
\left\{t_q\geq\underline{\tau_q},\ 
\widehat\mu_i^q\geq\widehat\mu_1^q\right\}
\]
is contained in the event that the empirical mean of arm $i$ exceeds
that of arm $1$ at some common sample size $t_q\geq\underline{\tau_q}$.  Hence
Lemma~\ref{lem:adaptiveHoeffding} gives
\[
\mathbb P\left(
 t_q\geq\underline{\tau_q},\ 
 \widehat\mu_i^q\geq\widehat\mu_1^q
 \mid\mathcal F_q, \mathcal{E}_q
\right)
\leq \exp(-\underline{\tau_q}\Delta_i^2).
\]
It follows that
\[
\mathbb E\left[N_q\mathbf 1\{t_q\geq\underline{\tau_q}\}
\,\middle|\,\mathcal F_q, \mathcal{E}_q \right]
\leq |S'^{(q)}|\exp(-\underline{\tau_q}\Delta_{i_q}^2),
\]
and therefore
\begin{equation}\label{eq:phase-elimination-stochastic}
\mathbb P\left(\overline{\mathcal{E}}_{q+1}
\,\middle|\,\mathcal F_q, \mathcal{E}_q,\mathcal{A}_q\right)
\leq
\exp\left(-18\underline{\tau_q}\Delta_{2n_q}^2\right)
+3\exp(-\underline{\tau_q}\Delta_{i_q}^2)\leq 4\exp(-\underline{\tau_q}\Delta_{i_q}^2).
\end{equation}

It remains to deal with the case $n_q=2$.
When $n_q=2$, there is only one competing arm.  Apply the adaptive
Hoeffding lemma directly, which also yields:
\begin{align*}
\mathbb P\left(\overline{\mathcal{E}}_{q+1}
\,\middle|\,\mathcal F_q, \mathcal{E}_q,\mathcal{A}_q\right)\leq
4\exp(-\underline{\tau_q}\Delta_{i_q}^2).
\end{align*}

\end{proof}

\subsection{Proof of the Helper Lemmas}

\begin{proof}[Proof of Lemma \ref{lem:9.1revisited}]

Fix a time $t$. If
\[
    |C_t(\epsilon)|\le g(\epsilon)-k,
\]
then at least $k$ of the $g(\epsilon)$ $\epsilon$-good arms are not selected. Since these $g(\epsilon)$ arms are contained in the first $\ell$ arms, the output contains at most $\ell-k$ arms from $[\ell]$. Because the output has size $\frac{K}{2}$, it must contain at least
\[
    \frac{K}{2}-(\ell-k)=\frac{K}{2}-\ell+k
\]
arms from $\{\ell+1,\dots,K\}$.

Thus, on the event $|C_t(\epsilon)|\le g(\epsilon)-k$, there exist sets
\[
    A\subset[g(\epsilon)],   \qquad |A|=k,
\]
and
\[
    B\subset\{\ell+1,\dots,K\},\qquad |B|=\frac{K}{2}-\ell+k,
\]
such that all arms in $A$ are excluded from the output and all arms in $B$ are included in the output. Since the output consists of the $\frac{K}{2}$ arms with largest empirical means, for every $i\in A$ and every $j\in B$,
\[
    \widehat\mu_{i,t}\le \widehat\mu_{j,t}.
\]
Moreover, for such $i,j$,
\[
    \mu_i-\mu_j
    \ge
    \mu_{g(\epsilon)}-\mu_{\ell+1}
    =
    d_\ell.
\]
Therefore, for each pair $(i,j)\in A\times B$, the inequality
$\widehat\mu_{i,t}\le \widehat\mu_{j,t}$ implies
\[
    \widehat\mu_{i}(t)\le \mu_i-\frac{d_\ell}{2}
    \quad\text{or}\quad
    \widehat\mu_{j}(t)\ge \mu_j+\frac{d_\ell}{2}.
\]
Indeed, if both inequalities failed, then
\[
    \widehat\mu_{i}(t)
    >
    \mu_i-\frac{d_\ell}{2}
    \ge
    \mu_j+\frac{d_\ell}{2}
    >
    \widehat\mu_{j}(t),
\]
a contradiction.

By distributivity,
\[
    \bigcap_{i\in A,j\in B}
    \left\{
        \widehat\mu_{i,t}\le \mu_i-\frac{d_\ell}{2}
        \ \text{or}\
        \widehat\mu_{j,t}\ge \mu_j+\frac{d_\ell}{2}
    \right\}
\]
implies, 
\[
    \left\{
        \forall i\in A,\ 
        \widehat\mu_{i}(t)\le \mu_i-\frac{d_\ell}{2}
    \right\}
    \cup
    \left\{
        \forall j\in B,\ 
        \widehat\mu_{j,t}\ge \mu_j+\frac{d_\ell}{2}
    \right\}.
\]
Hence,
\[
\begin{aligned}
&\left\{
\exists t\ge \underline \tau:
|C_t(\epsilon)|\le g(\epsilon)-k
\right\}
\\
&\subseteq
\left\{
\exists A\subset[g(\epsilon)],\ |A|=k,\ 
\exists t\ge \underline \tau:
\forall i\in A,\ 
\widehat\mu_{i}(t)\le \mu_i-\frac{d_\ell}{2}
\right\}
\\
&\quad\cup
\left\{
\exists B\subset\{\ell+1,\dots,K\},\ |B|=\frac{K}{2}-\ell+k,\ 
\exists t\ge \underline \tau:
\forall j\in B,\ 
\widehat\mu_{j}(t)\ge \mu_j+\frac{d_\ell}{2}
\right\}.
\end{aligned}
\]
Fix a set $A\subseteq[K]$. For any $i \in A$, by Claim \ref{lem:anytimeHoeff},
\[
\mathbb{P}\left(\exists t \geq \underline{\tau} \text{ s.t. } \hat{\mu}_i(t)\leq \mu_i-\frac{d_\ell}{2}\right)\leq \exp\left(-\frac{\underline{\tau}d_\ell^2}{2}\right).
\]
By independence, 
\[
\mathbb{P}\left(\exists t \geq \underline{\tau} \text{ s.t. } \forall i \in A, \hat{\mu}_i(t)\leq \mu_i-\frac{d_\ell}{2}\right)\leq \exp\left(-\frac{\underline{\tau}|A|d_\ell^2}{2}\right).
\]
Similarly, for a fixed set $B$,
\[
\mathbb{P}\left(\exists t \geq \underline{\tau}  \text{ s.t. } \forall i \in B, \hat{\mu}_i(t)\geq \mu_i+\frac{d_\ell}{2}\right)\leq \exp\left(-\frac{\underline{\tau}|B|d_\ell^2}{2}\right).
\]
Taking a union bound over all possible sets $A$ and $B$, and applying the
anytime concentration inequalities above, yields
\[
\begin{aligned}
\mathbb{P}\left(
\exists t\ge  \underline{\tau} :
|C_t(\epsilon)|\le g(\epsilon)-k
\right)
&\le
\binom{g(\epsilon)}{k}
\exp\left(-k\frac{d^2_\ell\underline \tau}{2}\right)
\\
&\quad+
\binom{K-\ell}{\frac{K}{2}-\ell+k}
\exp\left(-(\frac{K}{2}-\ell+k)\frac{d^2_\ell\underline \tau}{2}\right).
\end{aligned}
\]

\end{proof}

\begin{proof}[Proof of Lemma \ref{lem:meandecay}] Define
\[
    U(t):=B_s\setminus B_s(t),
    \qquad
    V(t):=B_s(t)\setminus B_s .
\]
By definition,
\[
\sum_{i\in B_s}\mu_i-\sum_{i\in B_s(t)} \mu_i= \sum_{i\in U(t)}\mu_i-\sum_{i\in V(t)} \mu_i.
\]
Hence, $\mu_{B_t(s)}\leq \mu_B-\epsilon$ implies 
\[\sum_{i\in U(t)}\mu_i-\sum_{i\in V(t)} \mu_i\geq s \epsilon.\] Since $B_s(t)$ is the set with the largest empirical means and the arms in $V(t)$ are selected while the arms in $U(t)$ are not, this implies:
\[
\sum_{i\in V(t)}\hat{\mu}_i(t)-\sum_{i\in U(t)} \hat{\mu}_i(t)\geq 0.
\]

Putting these two equations together, we obtain:
\[
\sum_{i\in U(t)}(\mu_i-\hat{\mu}_i(t))+\sum_{i\in V(t)} (\hat{\mu}_i(t)-\mu_i)\geq s\epsilon,
\]

or, adding $B_s \setminus U(t)$ to each set:
\[
\sum_{i\in B_s}(\mu_i-\hat{\mu}_i(t))+\sum_{i\in V(t)\cup(B_s\setminus U(t))} (\hat{\mu}_i(t)-\mu_i)\geq s\epsilon.
\]
Therefore:
\[
\left\{\mu_{B_t(s)}\leq \mu_B-\epsilon \right\} \subseteq \left\{\exists T, |T|=s \text{ and }\sum_{i\in B_s}(\mu_i-\hat{\mu}_i(t))+\sum_{i\in T} (\hat{\mu}_i(t)-\mu_i)\geq s\epsilon\right\}.
\]
Now, for a fixed set $T$, we prove the following claim. 

\begin{claim}
For every $\epsilon\geq 0$ and $\underline{\tau}>0$,
\[
\mathbb{P}\left(
\exists t\geq \underline{\tau}:
\sum_{i\in B_s}\bigl(\mu_i-\widehat{\mu}_i(t)\bigr)
+
\sum_{i\in T}\bigl(\widehat{\mu}_i(t)-\mu_i\bigr)
\geq s\epsilon
\right)
\leq
\exp\left(-s\underline{\tau}\epsilon^2\right).
\]

\end{claim}

\begin{proof}

Let $\ell = s-|B_s\cap T|$. Now, we can apply Claim \ref{lem:anytimeHoeff} with $m=2\ell \underline{\tau}$ and $a =\frac{s\epsilon}{2\ell}$, and get:

\[
\mathbb{P}\left(
\exists t\geq \underline{\tau}:
\sum_{i\in B_s}\bigl(\mu_i-\widehat{\mu}_i(t)\bigr)
+
\sum_{i\in T}\bigl(\widehat{\mu}_i(t)-\mu_i\bigr)
\geq s\epsilon
\right)
\leq
\exp\left(-\frac{s^2}{\ell}\underline{\tau}\epsilon^2\right).
\]
The claim then follows from $\ell\leq s$.

\end{proof}

Taking a union bound over all $\binom Ks$ deterministic sets $T$ gives
\[
\mathbb P\left(
    \exists t\geq \underline \tau:
    \mu_{B_t(s)}\leq \mu_B-\epsilon
\right)
\leq
\binom Ks
\exp\left(
    -\underline t\,s\epsilon^2
\right).
\]
Finally, using
\[
    \binom Ks\leq \left(\frac{eK}{s}\right)^s
\]
yields
\[
\mathbb P\left(
    \exists t\geq \underline \tau:
    \mu_{B_t(s)}\leq \mu_B-\epsilon
\right)
\leq
\exp\left(
    s\log\frac{eK}{s}
    -
    \underline \tau\,s\epsilon^2
\right).
\]

\end{proof}

\section{Experimental design details}

In the following section, we provide additional details on our experimental design. 

\begin{figure}[t]
\centering
\includegraphics[width=0.95\linewidth]{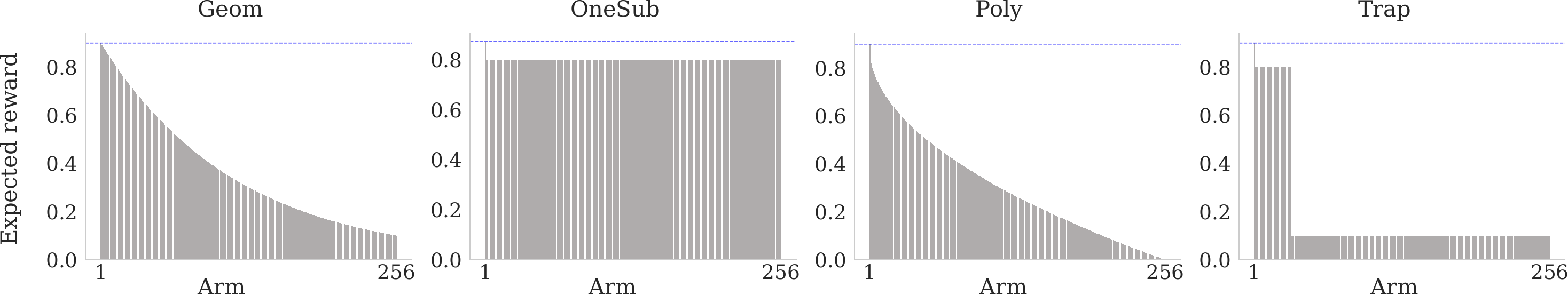}
\caption{Reward distributions in the four baseline configurations of the synthetic environments.}
\label{fig:synth_rewards}
\end{figure}

\textbf{Synthetic environments.} Figure~\ref{fig:synth_rewards} shows the reward structure of each synthetic environment. The rewards are defined as follows: \\
\textit{OneSub:} $\mu_1 = 0.9$, $\mu_i = 0.8$ for $i=2,\dots,K$, so all suboptimal arms share a common mean. \\
\textit{Trap:} $\mu_1 = 0.9$, $\mu_i = 0.8$ for $i=2,\dots,32$, and $\mu_i = 0.1$ for $i=33,\dots,K$, creating a small cluster of near-optimal competitors alongside a large set of clearly inferior, high-consumption arms. \\
\textit{Geom:} $\mu_i = 0.9\left(1/9\right)^{(i-1)/(K-1)}$ for $i=1,\dots,K$, yielding means decaying geometrically from $0.9$ to $0.1$. \\
\textit{Poly:} $\mu_1 = 0.9$ and $\mu_i = 0.9\left(1-\sqrt{i/K}\right)$ for $i\ge2$, producing polynomial decay with a higher density of near-optimal arms. 

\begin{figure}[t]
\centering
\includegraphics[width=0.95\linewidth]{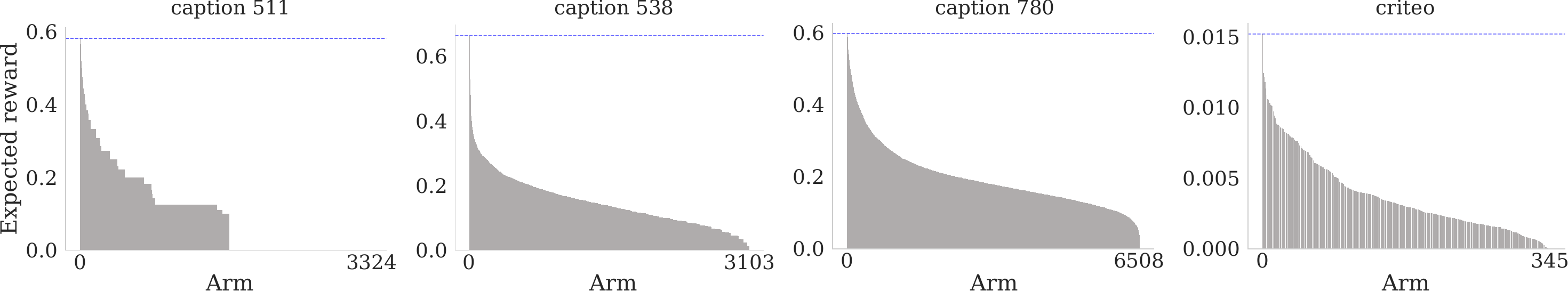}
\caption{Arm reward distributions in the four real-world environments: the three New Yorker Caption Contests (c511, c538, c780) and Criteo. Arms are sorted by mean reward in descending order.}
\label{fig:real_rewards}
\end{figure}
\textbf{Real-world environments.} Figure~\ref{fig:real_rewards} shows the arm reward distribution of each real-world environment.

\textit{New Yorker Caption Contest.}
We derive the first dataset from the New Yorker Caption Contest \cite{jain2020newyorker}, a widely used benchmark for preference learning and pure-exploration bandits. Each contest corresponds to a fixed cartoon and a finite set of submitted captions, which we treat as arms. Rewards are binary and reflect whether a caption is judged as \emph{somewhat funny} or \emph{funny} by human voters. For each arm, we estimate a ground-truth Bernoulli mean reward using the empirical proportion of positive votes aggregated across all voters, yielding an environment with human-generated noisy rewards and heterogeneous arm quality. We select contests 511, 538, and 780 (denoted c511, c538, c780), which induce three distinct empirical reward distributions and each contain a large number of arms (approximately 3,000--6,000), enabling evaluation with many arms. We use the top~1\% quantile arm as baseline, a regret budget of 1.6 times the number of arms in each contest, a maximum arm pull budget of 100,000, and 500 random seeds.

\textit{Criteo Attribution Modeling for Bidding.}
The second dataset is based on the Criteo Attribution Modeling for Bidding dataset \cite{DiemertMeynet2017}, which consists of large-scale logs from a real-world display advertising system. Each record corresponds to an \emph{ad impression}, i.e., a single instance in which an advertisement is displayed to a user. For each impression, the dataset records whether the displayed ad led to a conversion attributed to Criteo, the monetary value of that conversion, and the cost incurred by the advertiser to show the ad. The dataset is designed to support attribution and bidding decisions by providing both the revenue and cost components of advertising outcomes.

We aggregate impressions at the campaign level and treat each campaign as an arm. Campaigns with insufficient impression volume are filtered out to ensure stable estimation. For each campaign $i$, we denote by $\hat p_i$ the success (attribution) probability estimated from historical data and by $v_i$ the mean monetary value per successful attribution. When arm $i$ is pulled at round~$t$, an attribution outcome $A_t^{(i)} \sim \mathrm{Bernoulli}(\hat p_i)$ is generated, and the realized monetary reward is $r_t^{(i)} = v_i A_t^{(i)}$. The resulting expected rewards are very low, with a maximum arm mean of $0.015$, making this a challenging low-signal setting. Consumption is again defined via regret gaps relative to a baseline arm selected by a quantile of the empirical mean-reward distribution, yielding a natural interpretation of consumption as lost profit relative to a reference campaign. Unlike the caption dataset, this setting captures real economic trade-offs and monetary regret, providing a complementary evaluation setting.

This dataset provides a single real-world reward distribution, which we use to study the impact of different baseline choices: we select as baseline the top~5\% quantile arm, the top~1\% quantile arm, or the best arm, and evaluate performance under each. We sweep the regret budget from $0.06$ to $1.8$ times the number of arms, with a maximum arm pull budget of 100,000 and 500 random seeds.

\section{Per-Phase Arm Composition across Polynomial Reward Structures}
This section extends the per-phase composition analysis of the main paper to a
broader family of polynomial reward structures, varying the decay exponent to
test whether the effect of rationing on arm quality persists across reward
profiles of differing steepness.

Each figure corresponds to a reward family $\mathcal{P}(\beta)$ with
$\beta \in \{0.25, 0.5, 0.75, 1.0\}$, where arm rewards follow
$\mu_i = 0.9\cdot\left(1-(i/n)^\beta\right)$; larger $\beta$ yields a gentler
decay and therefore a higher density of near-optimal arms. In each figure, the
left panel shows the true arm reward distribution sorted in descending order.
The right panel shows, for each displayed phase, boxplots of the distribution
over 500 seeds of the mean reward among arms surviving to that phase, for four
\gls{acr:shprr} variants with $\alpha \in \{0.0, 0.2, 0.8, 1.0\}$. Boxes span
the interquartile range and the white bar marks the median. Significance
brackets indicate $t$-tests paired across seeds between each variant and the
$\alpha=0.0$ reference: $^{*}p<0.05$, $^{**}p<0.01$, $^{***}p<0.001$. Brackets
are shown only for statistically significant differences; their absence
indicates the comparison is not significant at the $p<0.05$ level. A star
($\star$) marks the variant with the highest mean reward at that phase.

\begin{figure}[t!]
\centering

\begin{subfigure}[t]{0.49\textwidth}
    \centering
    \includegraphics[width=\linewidth]{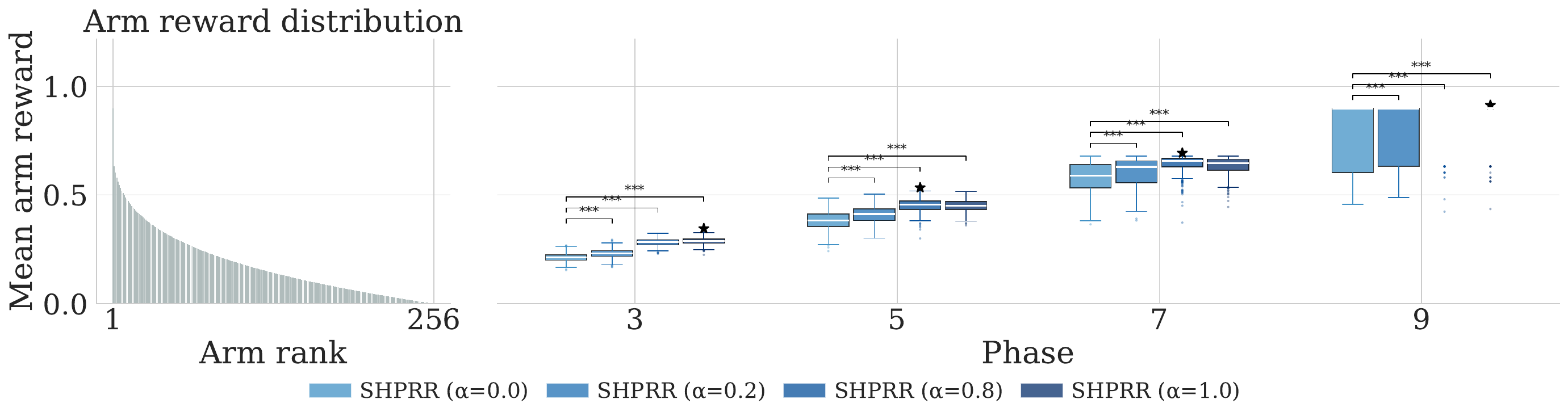}
    \caption{$\beta=0.25$}
    \label{fig:poly_variant_0p25}
\end{subfigure}
\hfill
\begin{subfigure}[t]{0.49\textwidth}
    \centering
    \includegraphics[width=\linewidth]{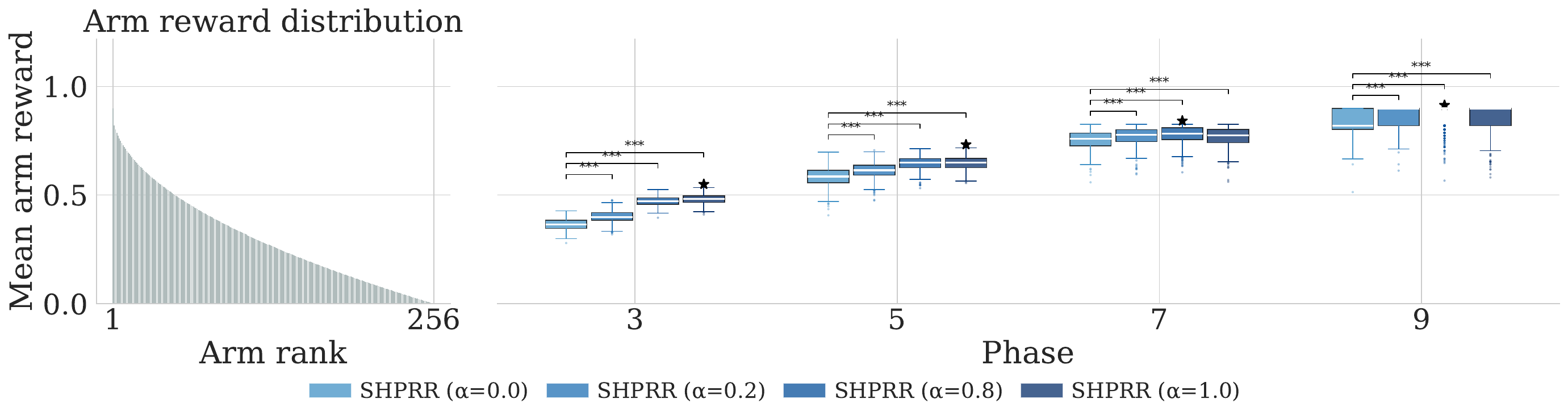}
    \caption{$\beta=0.5$}
    \label{fig:poly_variant_0p5}
\end{subfigure}

\vspace{0.5em}

\begin{subfigure}[t]{0.49\textwidth}
    \centering
    \includegraphics[width=\linewidth]{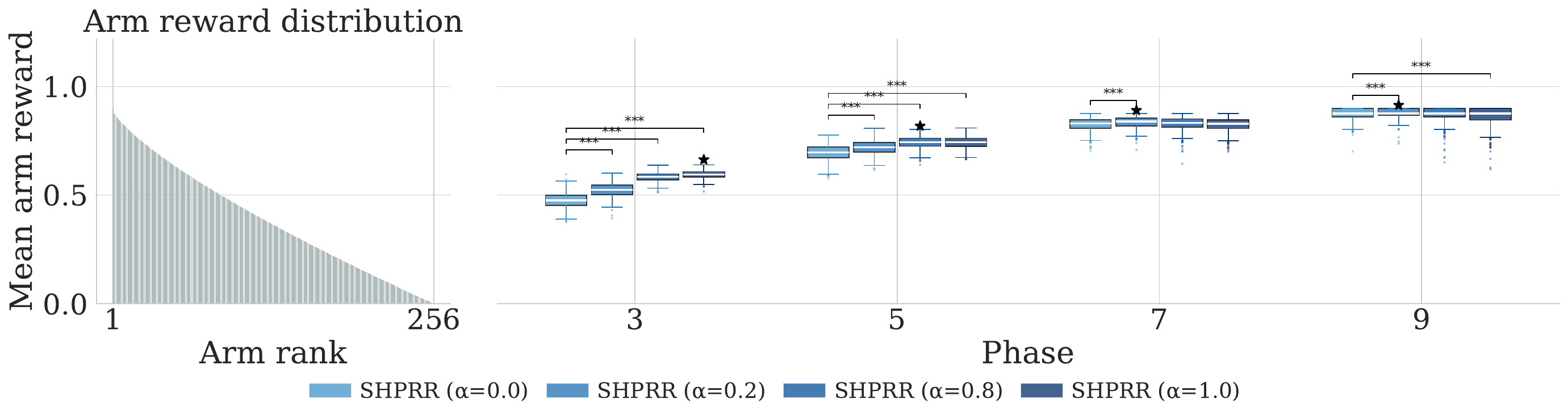}
    \caption{$\beta=0.75$}
    \label{fig:poly_variant_0p75}
\end{subfigure}
\hfill
\begin{subfigure}[t]{0.49\textwidth}
    \centering
    \includegraphics[width=\linewidth]{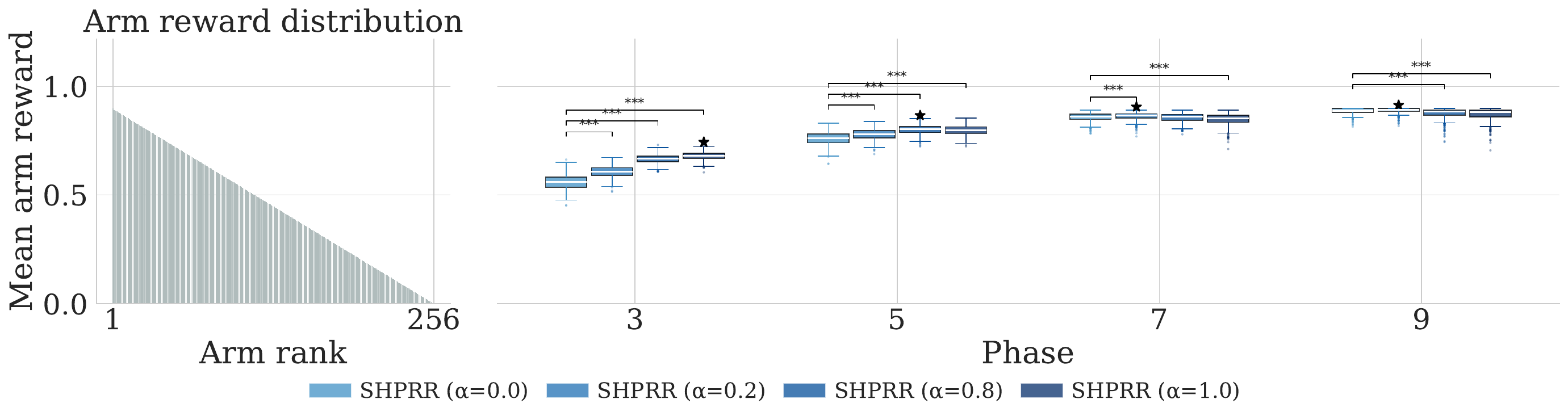}
    \caption{$\beta=1.0$}
    \label{fig:poly_variant_1p0}
\end{subfigure}

\caption{Phase-wise arm quality under a polynomial reward family, for $\beta \in \{0.25, 0.5, 0.75, 1.0\}$.}
\label{fig:poly_variant_all}
\end{figure}

\bibliographystyle{apalike}
\bibliography{aaai2027}